\PassOptionsToPackage{hyphens}{url}
\PassOptionsToPackage{table}{xcolor}
\documentclass{article}
\usepackage{iclr2027_conference,times}
\usepackage[utf8]{inputenc}
\usepackage[T1]{fontenc}
\usepackage{natbib}
\usepackage{hyperref}
\usepackage{url}
\usepackage{graphicx}
\usepackage{booktabs}
\usepackage{amsmath,amssymb,amsfonts}
\usepackage{xcolor}
\usepackage{adjustbox}
\usepackage{xspace}
\usepackage{caption}
\usepackage{subcaption}
\usepackage[many]{tcolorbox}
\usepackage[normalem]{ulem}
\usepackage{tabularx}
\usepackage{makecell}
\usepackage{tikz}
\usetikzlibrary{arrows.meta,positioning}

\fancypagestyle{evobenchfirstpage}{%
  \fancyhf{}%
  \fancyhead[L]{%
    \includegraphics[height=0.30in,pagebox=cropbox]{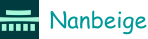}%
  }%
  \fancyhead[R]{%
    \includegraphics[height=0.30in]{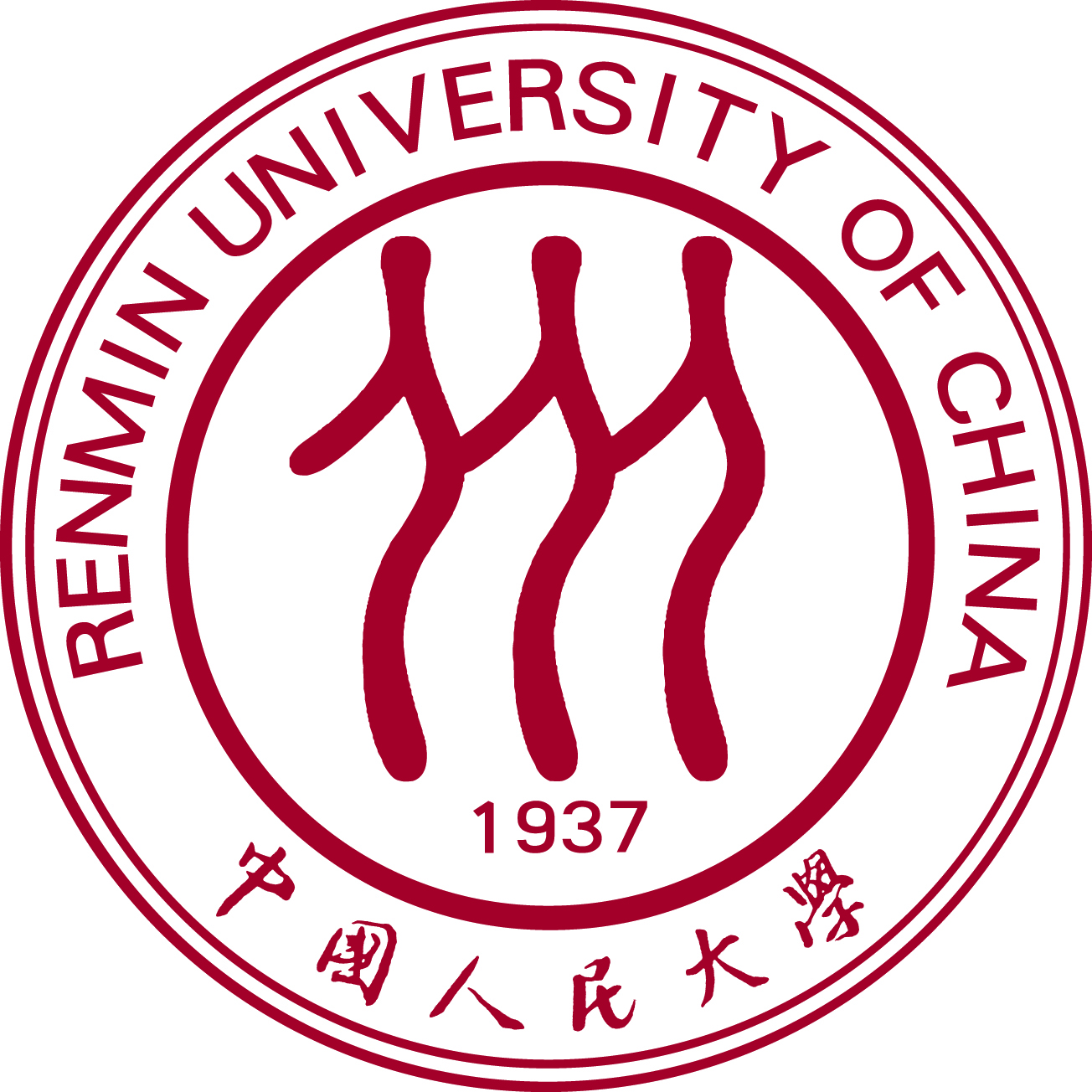}%
    \hspace{0.12in}%
    \includegraphics[height=0.29in]{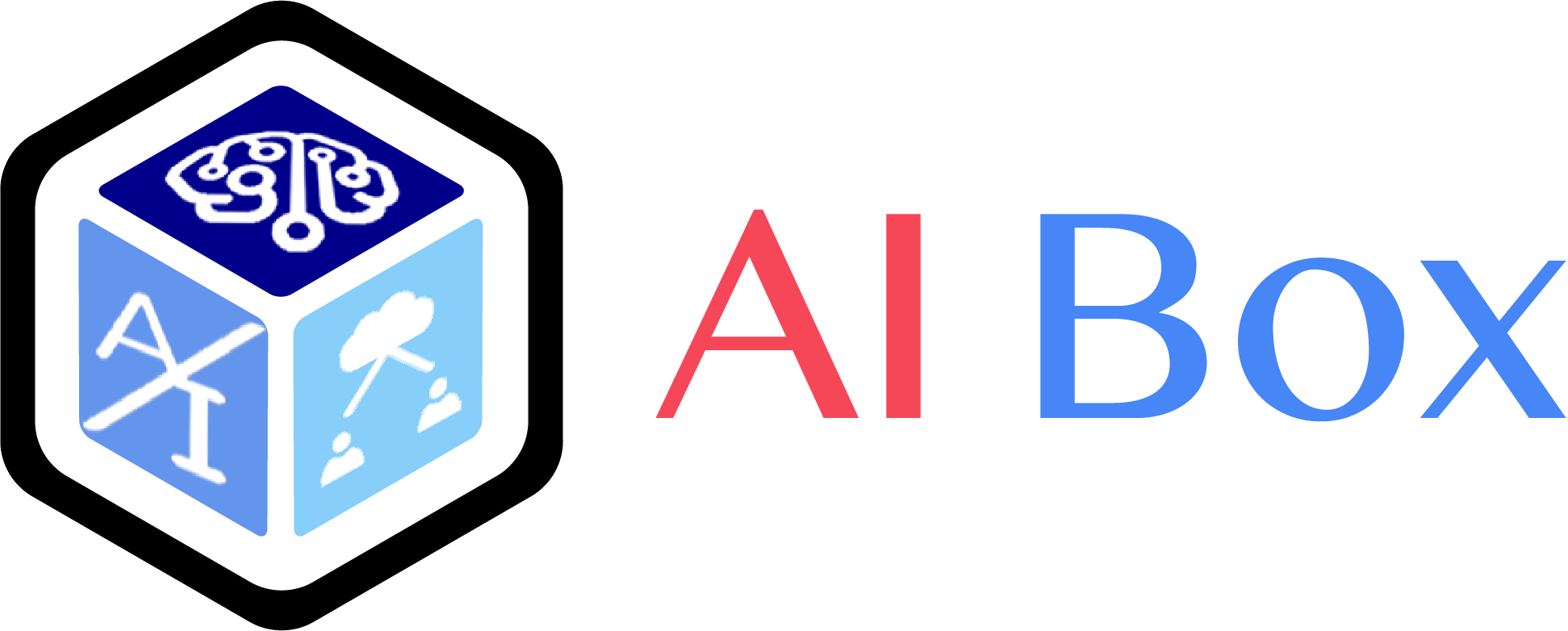}%
  }%
  \fancyfoot[C]{\thepage}%
  \renewcommand{\headrulewidth}{0.4pt}%
}

\newcommand{\frontmatterbutton}[3]{%
  \tcbox[
    on line,
    colback=black!4,
    colframe=black!16,
    boxrule=0.35pt,
    arc=1.7mm,
    left=1.8mm,
    right=1.8mm,
    top=0.8mm,
    bottom=0.8mm
  ]{\href{#1}{\textcolor{black}{\small #2\hspace{0.35em}#3}}}%
}

\newcommand{\huggingfaceicon}{%
  \tikz[baseline=-0.55ex,x=0.16em,y=0.16em,line cap=round,line join=round]{%
    \definecolor{hfbrand}{HTML}{FFD21E}%
    \fill[hfbrand] (0,0) circle (3.2);%
    \fill[black] (-1.1,0.8) circle (0.32);%
    \fill[black] (1.1,0.8) circle (0.32);%
    \draw[black,line width=0.35pt] (-1.35,-0.45) .. controls (0,-1.5) .. (1.35,-0.45);%
    \draw[hfbrand,line width=0.8pt] (-3,-0.45) .. controls (-4.2,-1.0) and (-4.3,-2.2) .. (-3.1,-2.55);%
    \draw[hfbrand,line width=0.8pt] (3,-0.45) .. controls (4.2,-1.0) and (4.3,-2.2) .. (3.1,-2.55);%
    \fill[hfbrand] (-4.05,-2.0) circle (0.52);%
    \fill[hfbrand] (4.05,-2.0) circle (0.52);%
  }%
}
\newcommand{\leaderboardicon}{%
  \tikz[baseline=-0.55ex,x=0.17em,y=0.17em,line cap=round,line join=round]{%
    \draw[line width=0.42pt] (0,0) circle (3);%
    \draw[line width=0.36pt] (-3,0) -- (3,0);%
    \draw[line width=0.36pt] (0,-3) .. controls (-1.25,-1.55) and (-1.25,1.55) .. (0,3);%
    \draw[line width=0.36pt] (0,-3) .. controls (1.25,-1.55) and (1.25,1.55) .. (0,3);%
    \draw[line width=0.36pt] (-2.55,-1.5) .. controls (0,-0.78) .. (2.55,-1.5);%
    \draw[line width=0.36pt] (-2.55,1.5) .. controls (0,0.78) .. (2.55,1.5);%
  }%
}

\hypersetup{
  colorlinks=true,
  linkcolor=black,
  citecolor=black,
  urlcolor=blue!50!black,
  pdftitle={Evo-Bench: Can Language Models Improve Agent Harness?},
  pdfauthor={Lisheng Huang, Chen Yang, Hao Zhou, Huatong Song, Zongchao Chen, Ran Le, Yang Song, Wayne Xin Zhao, Tao Zhang}
}

\newcommand{\paratitle}[1]{\vspace{1.5ex}\noindent\textbf{#1}}

\newcommand{\ignore}[1]{}

\title{Evo-Bench: Can Language Models Improve Agent Harness?}

\author{
  Lisheng Huang$^{1}$\thanks{Equal contribution.},
  Chen Yang$^{2}$\footnotemark[1],
  Hao Zhou$^{2}$,
  Huatong Song$^{1}$, \\
  \textbf{Zongchao Chen$^{2}$},
  \textbf{Ran Le$^{2}$},
  \textbf{Yang Song$^{2}$\thanks{Corresponding authors.}},
  \textbf{Wayne Xin Zhao$^{1}$\footnotemark[2]},
  \textbf{Tao Zhang$^{2}$} \\
  $^{1}$Gaoling School of Artificial Intelligence, Renmin University of China \\
  $^{2}$BOSS Zhipin, Beijing, China \\
  \texttt{huanglisheng@ruc.edu.cn},
  \texttt{batmanfly@gmail.com},
  \texttt{songyang@kanzhun.com}
}

\iclrfinalcopy

\begin{document}

\maketitle
\lhead{\footnotesize Evo-Bench: Can Language Models Improve Agent Harness?}
\fancyfoot[C]{\thepage}
\thispagestyle{evobenchfirstpage}

\vspace{-0.18in}
\noindent\makebox[\textwidth][c]{%
  \frontmatterbutton{http://evobench.org/}{\leaderboardicon}{Leaderboard}%
  \hspace{0.12in}%
  \frontmatterbutton{https://github.com/RUCAIBox/Evo-Bench}{\raisebox{-0.16em}{\includegraphics[height=0.95em]{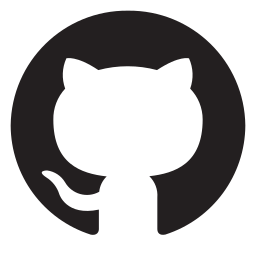}}}{Code}%
  \hspace{0.12in}%
  \frontmatterbutton{https://huggingface.co/datasets/RUC-AIBOX/Evo-Bench}{\huggingfaceicon}{Dataset}%
}\par
\vspace{0.04in}

\begin{abstract}
Large Language Models (LLMs) have driven rapid progress in autonomous agents, yet standard evaluations remain confined to static task solving. An emerging frontier is harness evolution---the agent's capacity to autonomously optimize its own operating harness. However, systematically benchmarking this capability remains challenging, as existing evaluations fail to isolate harness improvements from base model strength, prevent task-specific overfitting, or capture long-horizon iterative research. To address these challenges, we introduce \textbf{Evo-Bench}, the first benchmark designed to evaluate models' intrinsic harness-evolving capabilities across Search, Office, and General agent domains. To rigorously isolate this capability, Evo-Bench employs a novel harness-guided construction framework: it leverages auxiliary-task evolution to identify tasks genuinely sensitive to framework improvements, followed by sensitivity-aware stratified splitting to ensure robust cross-suite generalization. Extensive evaluations across nine frontier and open-weight models reveal that top models achieve massive absolute gains reaching 16.6 points, closely approaching state-of-the-art human-engineered baselines. Crucially, while autonomous evolution outpeforms artificial harness in General tasks and excels in Search tasks, it struggles in Office tasks that demand highly specific processing workflows. Furthermore, our analysis exposes critical temporal anomalies like early saturation, while demonstrating that the synthesized harnesses act as highly transferable reasoning structures, consistently boosting diverse policy models.
\end{abstract}


\section{Introduction}

Recent breakthroughs in large language models (LLMs) have remarkably elevated the capacity of agentic systems to execute complex, long-horizon tasks at an end-to-end, system-level scale~\cite{deepseekv4,glm5,anthropic2026opus,anthropic2026fable}. These gains arise not only from improvements in the underlying models, but also from carefully engineered agent harnesses that structure, coordinate, and constrain agent behavior, as exemplified by Claude Code~\cite{claudecode} and Codex~\cite{codex}.
In this context, an increasing body of work therefore seeks to explore  the potential for increasingly capable agentic systems to achieve self-improvement, including autonomous scientific research~\cite{alphaevolve}, automated model post-training~\cite{posttrainbench}, and autonomous harness evolution~\cite{dgm,metaharness,harnessx}. Among these lines of research, harness evolution is generally considered the first step toward achieving self-improvement~\cite{weng2026harness}.
Consequently, a pivotal question emerges for the foundational models: \emph{can language models truly improve agent harnesses, and how can we systematically benchmark this evolutionary capability?}

Recent work has explored harness evolution on two key fronts: methodologically, progressing from prompt search to end-to-end executable scaffold refactoring; and evaluatively, benchmarking evolution methods~\cite{seaeval,seagym} and agent development~\cite{mac} while analyzing underlying harness benefits~\cite{seaeval,seagym}. Collectively, these studies demonstrate how harnesses evolve and what constitutes success. However, systematically benchmarking an LLM's intrinsic capacity for harness evolution introduces unique, unaddressed obstacles.

Specifically, we identify three key challenges.
First, \emph{harness sensitivity}: benchmark task performance must be responsive to harness improvements rather than dominated by base model strength.
Second, \emph{cross-split generalization}: validation and evaluation splits must exhibit aligned responsiveness to prevent task-specific overfitting~\citep{rethinkingharnesseval}.
Third, \emph{long-horizon evolution}: models must sustain multi-round iterative refinement---diagnosing failures, formulating hypotheses, and progressively updating code.
Unaddressed, these challenges conflate true harness-evolving capability with model strength or task-specific overfitting.

To address these challenges, we introduce \textbf{Evo-Bench}, the first benchmark evaluating models' intrinsic harness-evolving capability. Evo-Bench adopts a controlled, long-horizon research setting with a fixed policy model across three domains (Search, Office, and General Agent tasks) over five established benchmarks~\cite{browsecomp,hle,gdpval,apexagents,claweval}. To rigorously construct this benchmark, we propose a novel \emph{harness-guided benchmark construction framework}. Built upon this framework, Evo-Bench incorporates three core designs: (1) \emph{harness-sensitive task construction}, identifying tasks responsive to framework improvements; (2) \emph{sensitivity-aware stratified splitting}, ensuring strict distributional consistency and disjoint validation and evaluation suites; and (3) \emph{controlled long-horizon evolution}, enabling sustained optimization of agent harnesses.

In our main evaluation across 9 frontier and open-weight models on \textbf{Evo-Bench}, top evolvers like GPT-5.6 Sol and Claude Opus 4.8 achieve substantial gains of 16.6 and 16.1 over the seed harness, respectively, closely approaching the human-engineered baseline of 47.5. Our analysis yields three key insights:
(1) \emph{evolutionary behavior exhibits early saturation}, with models rapidly discovering high-quality structures before introducing detrimental modifications in later rounds;
(2) \emph{gains are highly domain-dependent}, as evolvers effectively replicate web navigation in Search for massive gains and can even surpass manual engineering in General tasks, but struggle in Office tasks that demand highly specific processing workflows; and
(3) \emph{synthesized harnesses act as transferable reasoning structures}, demonstrating cross-policy robustness by driving consistent gains across diverse policy models from Qwen, DeepSeek, and GLM.

To summarize, our contributions are three-fold:

\begin{itemize}
\item We introduce \textbf{Evo-Bench}, the first benchmark evaluating LLMs' intrinsic harness-evolving capability---their ability to autonomously refine executable code harnesses across diverse domains beyond static task solving.

\item We propose a harness-guided benchmark construction framework that first induces diverse harnesses through auxiliary-task evolution and then constructs the final benchmark via harness-guided task selection.

\item We provide a systematic scientific account of harness evolution across 9 frontier models, characterizing substantial gains reaching 16.6 points, temporal anomalies like early saturation, and the robust transferability of evolved structures across distinct policy architectures.
\end{itemize}



\section{Related Work}


\paratitle{Automated Harness Engineering.} 
Recently, agentic self-improvement has evolved from prompt tuning to workflow optimization, and recently, to the automated refinement of the \emph{executable harness code} that orchestrates reasoning, tool use, and memory~\cite{weng2026harness}. While early prompt-level~\cite{promptbreeder,gepa,ace} and workflow-level methods~\cite{adas,aflow} operate on constrained design spaces, recent harness-level engineering leverages diverse mechanisms. These include self-referential scaffold refinement~\cite{stop,godelagent,dgm,selfharness}, evolutionary archive search over executable code~\cite{metaharness,vero}, and trajectory-guided optimization~\cite{harnessx}, alongside recent joint harness--weight co-optimization~\cite{sia}. 
Crucially, rather than proposing a new optimization algorithm, our work provides a complementary orthogonal perspective: we fix the policy model and research protocol to systematically benchmark the intrinsic capacity of frontier foundation models to act as long-horizon harness engineers across heterogeneous domains.


\paratitle{Agent Benchmarks.} 
Existing agent benchmarks predominantly measure task execution within static, fixed harnesses~\cite{swebench,osworld,taubench,browsecomp,gdpval}, whereas AI-R\&D benchmarks evaluate capabilities in optimizing model weights or training pipelines~\cite{mlagentbench,mlebench,rebench,paperbench,posttrainbench}. While the Meta-Agent Challenge evaluates the from-scratch generation of task-specific artifacts~\cite{mac}, an emerging class of evaluations focuses on self-evolutionary dynamics, tracking stability~\cite{seaeval}, distinguishing generation from utilization~\cite{harnessupdating}, assessing capability transfer~\cite{evoagentbench}, or formalizing diagnostic simulators~\cite{seagym}. 
Differing from these frameworks, \textbf{Evo-Bench} shifts the evaluation paradigm. Instead of evaluating task-specific artifacts or unconstrained self-evolution, we explicitly benchmark foundation models as long-horizon research agents tasked with evolving a single, shared general-purpose harness, validated under a rigorous, sensitivity-calibrated task split.

\section{Evo-Bench}
\label{sec:benchmark}

Evo-Bench is a benchmark for evaluating large language models' \emph{harness-evolving capability}: 
the ability to conduct long-horizon, code-centric iterative improvement of an executable agent harness. Given a set of validation tasks, models are expected to diagnose failure patterns, formulate improvement hypotheses, and revise the harness implementation to achieve sustained performance gains.

\begin{figure}[!t]
    \centering
    \includegraphics[width=0.84\linewidth]{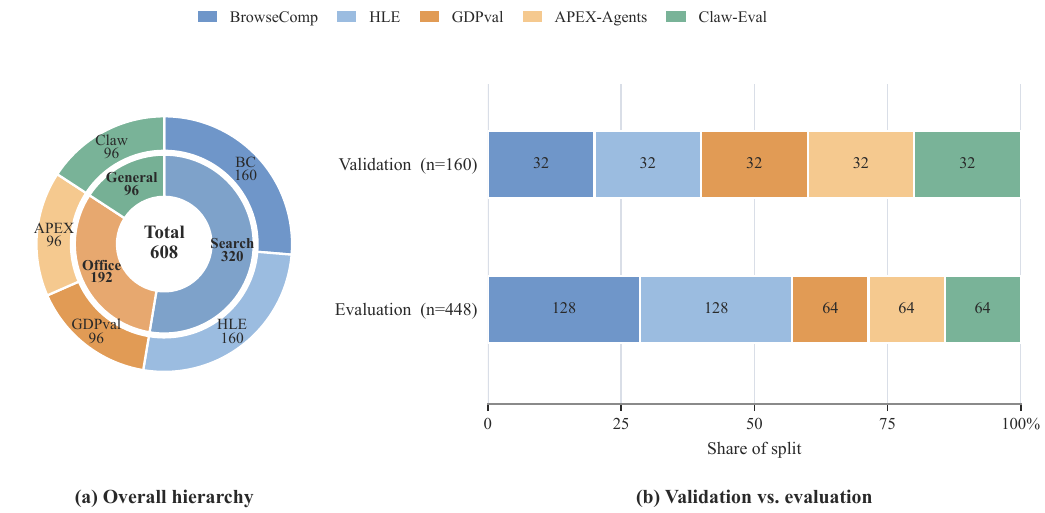}
    \caption{Composition of Evo-Bench.  The left panel summarizes the
    three-domain hierarchy and source-benchmark allocation; the right panel
    compares the validation and evaluation splits.}
    \label{fig:dataset-composition}
\end{figure}



\begin{figure}[!t]
    \centering
    \includegraphics[width=0.92\linewidth]{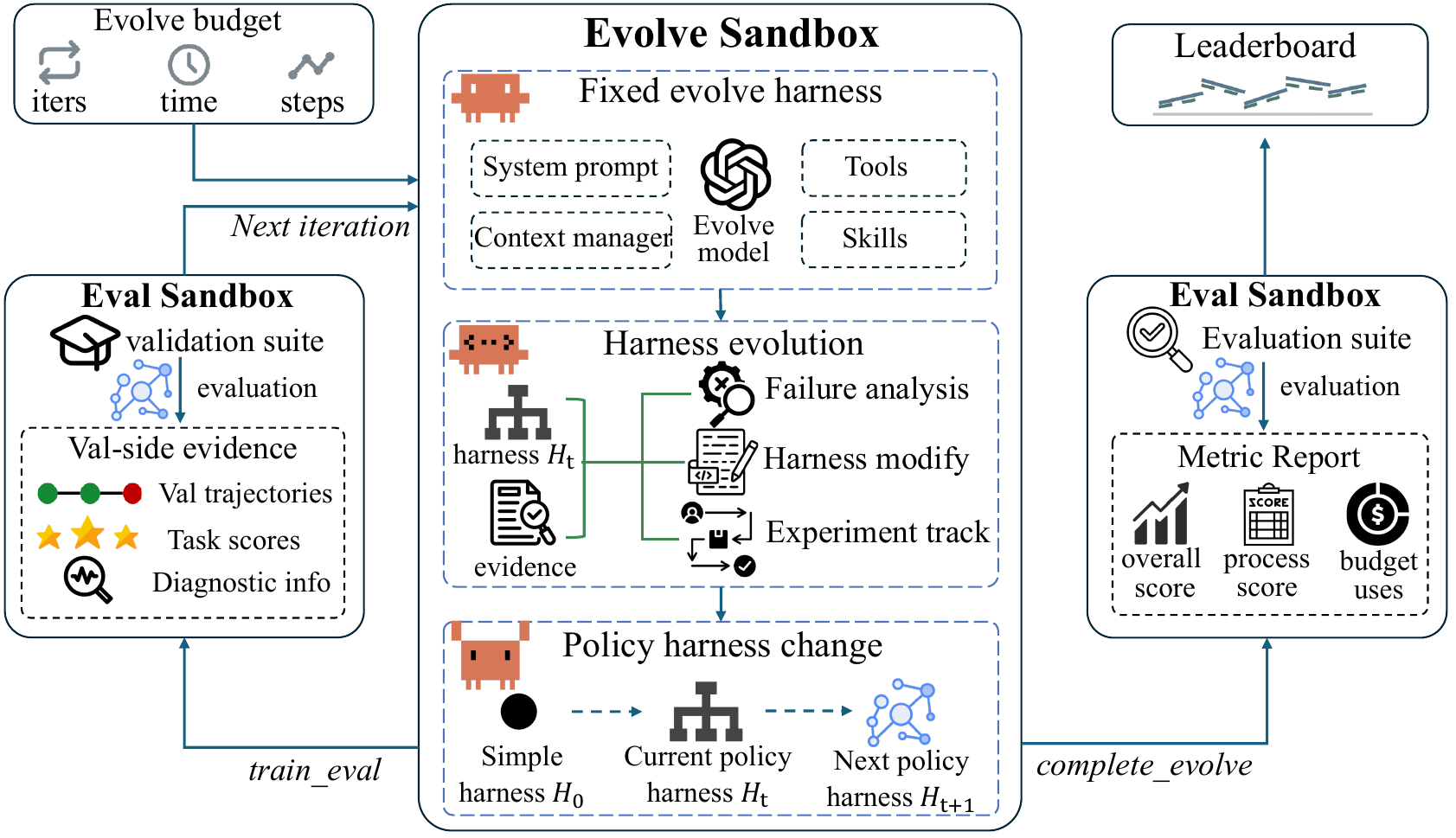}
    \caption{Overview of the Evo-Bench evaluation pipeline.}
    \label{fig:evaluation}
\end{figure}

\paratitle{Task Formulation.}
As shown in Figure~\ref{fig:evaluation}, each Evo-Bench run consists of a fixed policy model $\pi$ and an evolver $E$ that iteratively improves the policy harness.
At evolution iteration $t$, the policy agent is $A_t^{\mathrm{task}}=(\pi,H_t)$, where the editable policy harness $H_t$ governs how $\pi$ interacts with target tasks.
The evolver is $A^{\mathrm{evo}}=(E,\mathcal{H}_{\mathrm{evo}})$ and operates through a separate, fixed evolve harness $\mathcal{H}_{\mathrm{evo}}$. 
Let $\mathcal{E}_t^{\mathrm{val}}:=\bigl((H_i,j_i^{\mathrm{val}},O_i^{\mathrm{val}})\bigr)_{i<t}$ denote the cumulative validation-side evidence available at iteration $t$, where $j_i^{\mathrm{val}}$ is the aggregate validation score and $O_i^{\mathrm{val}}$ contains the task-level outcomes, policy trajectories, and diagnostic feedback from evaluation $i$.
Starting from the common seed harness $H_0$, the evolver inspects the accumulated validation-side evidence, edits the current harness $H_t$, and requests a formal evaluation of the resulting revision on the visible validation suite at each iteration. 
The evolution process is constrained by a fixed resource budget $\mathbf{b}=(b^{\mathrm{iter}},b^{\mathrm{time}},b^{\mathrm{steps}})$, which bounds the number of iterations, wall-clock time, and evolver steps. When the run terminates, the final revision $H_T$ is frozen and evaluated on a disjoint held-out evaluation suite.


\paratitle{Task Domains and Composition.}
\label{sec:benchmark-composition}
Evo-Bench covers three representative agent domains: search, office, and general agent tasks, evaluating harness evolution across diverse interaction patterns and tool-use requirements.
To instantiate these domains, we select five  established and challenging benchmarks: BrowseComp~\citep{browsecomp} and HLE~\citep{hle} for search, GDPval~\citep{gdpval} and APEX-Agents~\citep{apexagents} for office, and Claw-Eval~\citep{claweval} for general agent tasks. 
Based on these benchmarks, we construct a 160-task visible validation suite $\mathcal{D}_{\mathrm{val}}$ and a disjoint 448-task evaluation suite $\mathcal{D}_{\mathrm{eval}}$. 
Each source benchmark contributes 32 validation tasks, while the evaluation suite contains 128 BrowseComp, 128 HLE, 64 GDPval, 64 APEX-Agents, and 64 Claw-Eval tasks.
During evolution, the evolver can request validation evaluations only on $\mathcal{D}_{\mathrm{val}}$, while $\mathcal{D}_{\mathrm{eval}}$ is reserved for final evaluation after the harness is frozen.
Figure~\ref{fig:dataset-composition} summarizes the domain hierarchy and the
per-source allocation across the two splits.


\paratitle{Evolve Harness.}
\label{sec:evolve-harness}
The evolve harness operates as a fixed agent loop explicitly designed to support long-horizon autonomous work. Its core components, including the system prompt, execution tools, and context manager, are inspired by the design of Claude Code. Building upon this foundation, we integrate specialized skills for the evolution process, such as trajectory analysis and experiment tracking. We implement this custom harness natively within Evo-Bench to enable a controlled and transparent evaluation protocol.

\paratitle{Policy Harness.}
\label{sec:policy-harness}
The policy harness is the evolving artifact optimized by the evolver. Its initial version $H_0$ is a minimal CodeAct~\cite{wang2024executable} loop with only a shell-execution tool and a final-answer completion tool. During validation evaluation on $\mathcal{D}_{\mathrm{val}}$, the fixed policy model executes tasks in a separate sandbox through the current harness $H_t$.
The evolver aims to transform this initial harness into a more capable, generalizable harness that boosts performance across all three domains.

\paratitle{Evaluation Metrics.}
\label{sec:benchmar_metrics}
We evaluate harness-evolving capability from two complementary perspectives: final generalization performance and evolutionary progress.
For a given suite $\mathcal{D}$, let $\mathcal{S}(\pi, H; \mathcal{D})$ denote the aggregate score achieved by the policy model with harness $H$, computed from the source benchmarks' native scorer~(defined in Table~\ref{tab:task-selection-stats}) and aggregated across target domains. 

To measure the final performance of the evolved harness, we define \textbf{Overall Score} on the unseen evaluation suite:
\[
\mathrm{Overall}(E)
:=\mathcal{S}(\pi,H_T;\mathcal{D}_{\mathrm{eval}}).
\]

To measure the evolution progress, we define \textbf{Anytime Validation Score} as the average best-so-far validation performance over the evolution budget. At iteration $t$, the best validation score achieved up to and including iteration $t$ is:
\[
S_t^*=\max_{i\leq t}\mathcal{S}(\pi,H_i;\mathcal{D}_{\mathrm{val}}).
\]

The Anytime Validation Score is then defined as:
\[
\mathrm{AnytimeVal}(E)
:=\frac{1}{b^{\mathrm{iter}}}
\sum_{t=1}^{b^{\mathrm{iter}}}S_t^* .
\]

If a run terminates early, its final best-so-far validation score is carried forward for the remaining budget.

\begin{table}[htbp]
\centering
\caption{Statistics of the harness-guided task selection process. Tasks with non-positive sensitivity ($\mathrm{Sens}(x) \le 0$) are filtered out prior to sensitivity-aware stratified selection. \textbf{Metrics}: $\mathrm{Sens}(x)$ denotes harness sensitivity (Pearson correlation); $\mathrm{Perf}(x)$ denotes average performance across the auxiliary harness set $\mathcal{H}_{\mathrm{aux}}$. ``LJ'' represents LLM-as-a-Judge.}
\label{tab:task-selection-stats}
\resizebox{\textwidth}{!}{%
\begin{tabular}{l ll cc cc cc cc}
\toprule
& \multicolumn{2}{c}{\textbf{Evaluation Protocol}} & \multicolumn{2}{c}{\textbf{Candidate Filtering}} & \multicolumn{2}{c}{\textbf{Final Allocation}} & \multicolumn{2}{c}{\textbf{$\mathrm{Sens}(x)$}} & \multicolumn{2}{c}{\textbf{$\mathrm{Perf}(x)$}} \\
\cmidrule(lr){2-3} \cmidrule(lr){4-5} \cmidrule(lr){6-7} \cmidrule(lr){8-9} \cmidrule(lr){10-11}
{Source Dataset} & Evaluator & Metric & Candidates & $\mathrm{Sens}(x) \le 0$  & Validation & Evaluation & Mean & Median & All & Selected \\
\midrule
APEX-Agents & Rubric LJ & Pass@1 & 421 & 133 & 32 & 64  & 0.377 & 0.423 & 0.389 & 0.224 \\
BrowseComp  & LJ        & Pass@1 & 768 & 47  & 32 & 128 & 0.256 & 0.236 & 0.444 & 0.271 \\
Claw-Eval   & LJ+Rule   & Pass\^{}$3$ & 157 & 60  & 32 & 64  & 0.386 & 0.317 & 0.839 & 0.747 \\
GDPval      & Rubric LJ & Mean    & 215 & 95  & 32 & 64  & 0.368 & 0.336 & 0.797 & 0.495 \\
HLE         & LJ        & Pass@1 & 768 & 62  & 32 & 128 & 0.313 & 0.300 & 0.333 & 0.251 \\
\bottomrule
\end{tabular}
}
\end{table}

\section{Benchmark Construction}

\begin{figure}[htbp]
    \centering
    \includegraphics[width=1.0\linewidth]{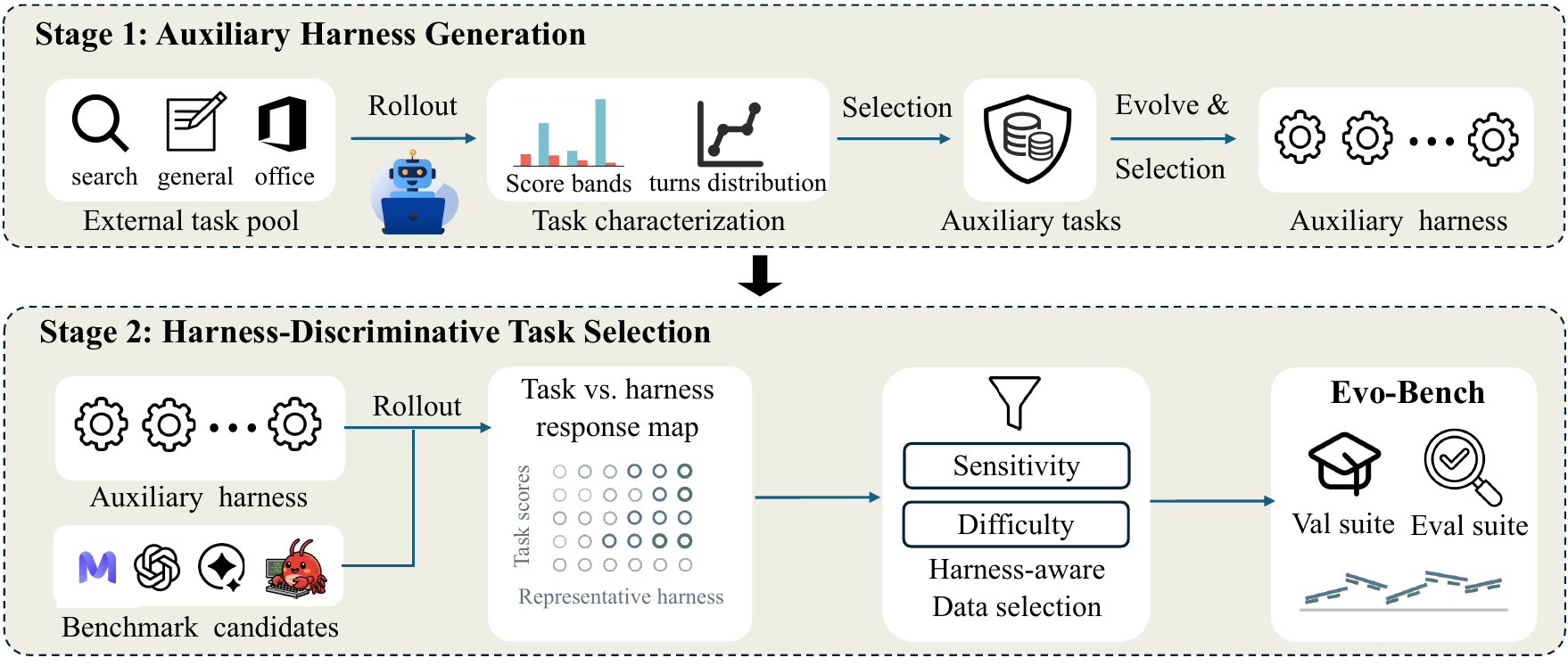}
    \caption{Overview of Two-Stage Harness-Guided Benchmark Construction Framework.}
    \label{fig:benchmark_construction}
\end{figure}

To ensure that validation-side optimization reliably predicts held-out evaluation performance, benchmark tasks should satisfy two key properties: \emph{Harness Sensitivity}, where task performance is sensitive to harness improvements while controlling for the underlying model, and \emph{Cross-Suite Alignment}, where validation and evaluation suites exhibit consistent performance to harness variation. To this end, we design a Two-Stage Harness-Guided Benchmark Construction Framework. As shown in Figure~\ref{fig:benchmark_construction},
we first leverage auxiliary tasks to conduct controlled evolution experiments and generate a diverse set of auxiliary evolved harnesses, and then use these harnesses to characterize task-level harness sensitivity and difficulty for constructing the final benchmark task suites.

\label{sec:task-construction}
\subsection{Stage 1: Auxiliary Harness Generation}
\label{sec:calibration-tasks}

\paratitle{Auxiliary Tasks Collection.}
We collect auxiliary tasks from diverse sources that are corpus- and instance-level disjoint from the five source benchmark datasets of Evo-Bench, including MiroRL~\cite{2025mirorl}, RedSearcher~\cite{redsearcher2026}, Auto-ClawEval~\cite{li2026clawenvkitautomaticenvironmentgeneration}, and internally constructed datasets. 
We then evaluate these tasks with DeepSeek-V4-Flash~\cite{deepseekv4} through multiple independent rollouts and select tasks with lower average scores and longer interaction trajectories, which indicate greater room for harness improvements and stronger signals for harness evolution. This process results in a 320-task auxiliary set, comprising 128 search tasks, 128 office tasks, and 64 general agent tasks.

\paratitle{Auxiliary Harness Evolution.}
After constructing the auxiliary task set, we run independent evolution on it with multiple frontier models, collecting resulting harnesses throughout the process. We then deduplicate and select a representative subset to ensure sufficient diversity and coverage.

Specifically, we conduct four independent evolution experiments using four frontier models: GLM-5.2~\cite{glm52}, Claude-Opus-4.8~\cite{anthropic2026opus}, Claude-Sonnet-5~\cite{anthropic2026sonnet}, and GPT-5.6-Sol~\cite{gpt56}. Each experiment follows the same policy model, evolution protocol, and resource budget as the benchmark evaluation. We collect 73 formally evaluated harness variants and apply a deterministic diversity-aware selection procedure to identify 12 representative harnesses that maximize diversity across harness capabilities, tool orchestration, program structure, and auxiliary-task performance. We denote this selected set as $\mathcal{H}_{\mathrm{aux}}=\{h_1,\ldots,h_K\}$, where $K=12$.

\subsection{Stage 2: Harness-Guided Task Selection}
\label{sec:harness-response-prior}

We evaluate candidate tasks under the selected harness set $\mathcal{H}_{\mathrm{aux}}$ to identify tasks that reliably reflect harness improvements. Specifically, we evaluate 2,329 candidate tasks collected from APEX-Agents, BrowseComp, Claw-Eval, GDPval, and HLE using the 12 selected harnesses. We exclude multimodal tasks to maintain a unified interface and downsample BrowseComp and HLE to control evaluation cost while preserving their original task distributions.

For task $x$ from the public benchmark $\mathcal{D}_{\mathrm{public}}$, let $m_h(x)$ denote its score under harness $h\in\mathcal{H}_{\mathrm{aux}}$. 
A straightforward criterion is to select tasks with high score variance across harnesses.
However, high variance only captures the magnitude of performance differences and does not indicate whether these differences reflect the relative quality of harnesses.
To more precisely measure whether a task consistently responds to harness quality, we compute the correlation between the task and overall harness quality, and combine this sensitivity measure with the average task score for task selection.

\paratitle{Harness Sensitivity and Task Difficulty.}
The score of an individual task is obtained directly from its execution result, while the overall quality of each harness is measured by its average performance across the remaining tasks. We compute the Pearson correlation coefficient between task-level scores and overall harness quality as the metric for task sensitivity. 
Specifically, we define two task-level metrics: harness sensitivity and average performance, as follows:

\begin{equation*}
\begin{aligned}
\mathrm{Sens}(x)
&=
\operatorname{corr}
\left(
\{m_h(x)\}_{h\in\mathcal{H}_{\mathrm{aux}}},
\{Q_h^{(-x)}\}_{h\in\mathcal{H}_{\mathrm{aux}}}
\right),\\
\mathrm{Perf}(x)
&=
\frac{1}{|\mathcal{H}_{\mathrm{aux}}|}
\sum_{h\in\mathcal{H}_{\mathrm{aux}}}m_h(x),
\end{aligned}
\end{equation*}
where $Q_h^{(-x)}$ denotes the leave-one-task-out average performance of harness $h$, computed over all tasks except $x$, and serves as a robust estimate of its overall quality.

Higher $\mathrm{Sens}(x)$ indicates that task $x$ consistently reflects the quality ranking of different harnesses, where stronger harnesses achieve better performance on the task. Therefore, tasks with higher sensitivity provide more reliable signals for evaluating harness evolution. In contrast, $\mathrm{Perf}(x)$ measures the average performance across harnesses, and we define task difficulty as $1-\mathrm{Perf}(x)$, where lower average performance indicates larger performance headroom.



\paratitle{Task Suite Construction.}
We first remove tasks with non-positive $\mathrm{Sens}(x)$, as they provide limited signal for distinguishing harness quality. Among the remaining tasks, we partition them into difficulty strata according to $1-\mathrm{Perf}(x)$ and select tasks with the highest $\mathrm{Sens}(x)$ within each stratum. This procedure yields a task set that preserves difficulty diversity while prioritizing tasks that are more responsive to harness evolution. Then we randomly split tasks within each stratum into validation and evaluation suites, ensuring that they follow the same difficulty distribution. Table~\ref{tab:task-selection-stats} reports the intermediate statistics of the task selection process.


\definecolor{BaselineBlue}{HTML}{E3F5FB}
\definecolor{EvoGainGreen}{RGB}{0,140,45}
\definecolor{EvoLossRed}{RGB}{190,20,32}
\definecolor{EvoNeutralGray}{RGB}{110,110,110}
\newlength{\EvoDeltaWidth}
\settowidth{\EvoDeltaWidth}{{\scriptsize +00.0}}

\newcommand{\gain}[1]{\,\textcolor{EvoGainGreen}{\scriptsize +#1}}
\newcommand{\loss}[1]{\,\textcolor{EvoLossRed}{\scriptsize -#1}}
\newcommand{\same}{\,\textcolor{EvoNeutralGray}{\scriptsize $\pm$0.0}}
\newcommand{\nodelta}{%
  \,\makebox[\EvoDeltaWidth][l]{}%
}

\begin{table*}[t]
\centering
\caption{
Evo-Bench harness-evolving capability leaderboard.
Methods are ranked by Overall score; the two baseline harnesses are excluded from ranking. \textbf{Bold} and \underline{underlined} denote the best and second-best raw scores for each metric, respectively. 
\textcolor{EvoGainGreen}{Green} and \textcolor{EvoLossRed}{red} values indicate absolute score changes relative to the CodeAct baseline; \textcolor{EvoNeutralGray}{gray} indicates no change.
}
\label{tab:main-leaderboard}
\small
\resizebox{\textwidth}{!}{%
\begin{tabular}{c l ccccc}
\toprule
\textbf{\rule{0pt}{2ex}Rank}
& \textbf{Model}
& \textbf{Search score}
& \textbf{Office score}
& \textbf{General score}
& \textbf{Overall score}
& \textbf{AnytimeVal} \\
\midrule
\rowcolor{BaselineBlue}
\multicolumn{7}{c}{\textit{Frontier models}} \\
\midrule
1 & GPT-5.6 Sol~\cite{gpt56}
  & 44.5\gain{32.8}
  & \underline{41.6}\gain{3.2~}
  & \textbf{59.4}\gain{11.0}
  & \textbf{46.3}\gain{16.6}
  & 50.1 \\
2 & Claude Opus 4.8~\cite{anthropic2026opus}
  & \textbf{46.5}\gain{34.8}
  & 39.7\gain{1.3~}
  & \underline{56.3}\gain{7.9~}
  & \underline{45.8}\gain{16.1}
  & \textbf{51.4} \\
3 & GLM-5.2~\cite{glm52}
  & \underline{45.4}\gain{33.7}
  & 39.2\gain{0.8~}
  & 48.4\same
  & 43.5\gain{13.8}
  & \underline{51.0} \\
4 & Qwen3.7-Max~\cite{qwen37}
  & 36.3\gain{24.6}
  & 37.8\loss{0.6~}
  & \textbf{59.4}\gain{11.0}
  & 41.5\gain{11.8}
  & 49.3 \\
5 & Minimax-M3~\cite{lai2026minimax}
  & 33.6\gain{21.9}
  & \textbf{41.7}\gain{3.3~}
  & \underline{56.3}\gain{7.9~}
  & 41.4\gain{11.7}
  & 49.0 \\
7 & Deepseek V4 Pro~\cite{deepseekv4}
  & 34.4\gain{22.7}
  & 39.1\gain{0.7~}
  & 48.4\same
  & 39.1\gain{9.4~}
  & 45.4 \\
8 & Kimi K2.7 Code~\cite{kimik27}
  & 34.5\gain{22.8}
  & 38.1\loss{0.3~}
  & 48.4\same
  & 38.7\gain{9.0~}
  & 43.4 \\
\midrule
\rowcolor{BaselineBlue}
\multicolumn{7}{c}{\textit{Open-weight models}} \\
\midrule
6 & Qwen3.6-27b~\cite{qwen3.6-27b}
  & 34.8\gain{23.1}
  & 38.8\gain{0.4}
  & 50.0\gain{1.6}
  & 39.4\gain{9.7}
  & 46.9 \\
9 & Gemma-4-31B~\cite{gemmateam2026gemma4}
  & 24.2\gain{12.5}
  & 40.4\gain{2.0}
  & 50.0\gain{1.6}
  & 35.9\gain{6.2}
  & 36.2 \\
\midrule
\rowcolor{BaselineBlue}
\multicolumn{7}{c}{\textit{Baseline variants}} \\
\midrule
-- & CodeAct
   & 11.7\nodelta
   & ~38.4\nodelta
   & ~48.4\nodelta
   & ~29.7\nodelta
   & -- \\
-- & Artificial Harness
   & 46.7\nodelta
   & ~43.9\nodelta
   & ~56.3\nodelta
   & ~47.5\nodelta
   & -- \\
\bottomrule
\end{tabular}
}
\end{table*}

\section{Experiments}
\label{sec:experiments}
\raggedbottom


In this section, we present the main results of Evo-Bench. We evaluate the harness-evolving capabilities of frontier models, analyze their evolve process, and further discuss under what conditions model-evolved harnesses can surpass the current state-of-the-art human-engineered frameworks.

\subsection{Experimental Setup}
\label{sec:exp-setup}

\paratitle{Baselines.}
We compare the evolved harnesses against two baselines: (1) the initial \textbf{CodeAct} seed harness, which serves as the starting point to quantify absolute evolutionary gains, and (2) \textbf{Artificial harness}, a composite of domain-specific human-engineered frameworks: MiroFlow~\cite{miromind2025mirothinker} for search, Stirrup~\cite{artificialanalysis2026stirrup} for office, and Claw-Eval~\cite{claweval} for general tasks.

\subsubsection{Evaluation protocol.}
Unless stated otherwise, all experiments use DeepSeek-V4-Flash as the fixed policy model and start from the same CodeAct seed policy harness $H_0$. The main experiments use a common budget of 20 iterations, 1,000 evolver steps, and 48 hours. Each policy rollout is capped at 300 steps and one hour. Search and office tasks use one rollout, whereas Claw-Eval use three rollouts, following its native three-trial $\mathrm{Pass}$\^{}$3$ metric. Tasks requiring LLM-based grading use Qwen3.7-Plus as the common judge. We run all experiments one time, and report domain-level scores, Overall Score, and Anytime Validation Score (AnytimeVal), as defined in the previous section. Costs are computed from recorded input, cached-input, and output tokens using provider prices as of July 10, 2026.

\subsubsection{Model configuration.}
We evaluate seven frontier models and two open-weight models on Evo-Bench across the following settings. 
All evolvers run in thinking mode with temperature $1.0$ and the largest context window supported by the model. For models exposing a reasoning-effort control, we use their maximum setting; for models that do not expose a configurable reasoning-effort level, we use their default setting. The shared DeepSeek-V4-Flash policy model uses maximum reasoning effort, temperature $1.0$, and a 256K-token context window. The Qwen3.7-Plus judge uses temperature $0.0$ and a 1M-token context window. 

\subsection{Main Results}

\paratitle{Overall performance.}
Table~\ref{tab:main-leaderboard} presents the overall leaderboard of all evaluated models. GPT-5.6-Sol and Claude Opus-4.8 lead with scores of 46.3 and 45.8 respectively, yielding massive absolute gains over the initial CodeAct baseline. This widespread positive delta validates that frontier LLMs possess a genuine capability to autonomously optimize executable harnesses. Examining evolutionary progress, Claude Opus-4.8 and GLM-5.2 achieve the highest Anytime Validation scores. This early peak suggests they rapidly evolve high-quality structures but frequently introduce detrimental modifications in subsequent iterations. Despite these gains, the top evolved harness still slightly underperforms domain-specific Artificial harness composite score of 47.5, leaving headroom for future research.

\paratitle{Per-domain performance.}
The evolutionary gains are highly uneven across different task domains. \textbf{Search} tasks are more amenable to optimization, with Claude Opus-4.8 gaining +34.8 to nearly match the Artificial harness, indicating evolvers easily synthesize missing web-navigation logic. Conversely, \textbf{Office} tasks remain stubborn. Most models show marginal improvements or slight regressions, failing to match the Artificial baseline likely due to the difficulty of discovering specialized workflows. Interestingly, for \textbf{General} tasks, top evolvers like GPT-5.6-Sol and Qwen3.7-Max strictly surpass the Artificial harness. This milestone demonstrates that autonomously evolved reasoning structures can indeed outstrip manually crafted solutions.

\begin{figure}[htbp]
    \centering
    \includegraphics[width=0.9\linewidth]{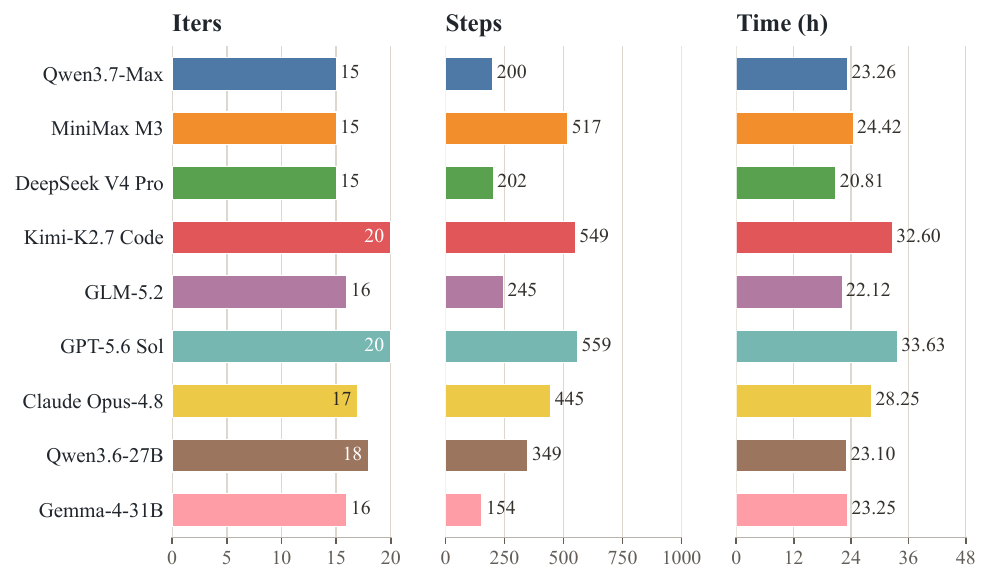}
    \caption{Budget usage of each evolver. Iters, steps and time report the number used relative to the maximum budgets of 20 iterations, 1,000 steps, and 48 hours, respectively.}
    \label{fig:budget_use}
\end{figure}

\paratitle{Budget use.}
Figure~\ref{fig:budget_use} reveals two distinct evolutionary behaviors: exhaustive exploration and early saturation. GPT-5.6-Sol and Kimi-K2.7-Code uniquely exhaust the maximum 20-iteration budget with the highest step counts and longest durations. This persistent exploration directly contributes to the top-tier overall performance of GPT-5.6-Sol. Conversely, most models terminate prematurely due to invalid code proposals or stagnant reasoning loops. For instance, Qwen3.7-Max and DeepSeek-V4-Pro halt at 15 iterations using barely 200 steps. However, budget exhaustion does not strictly dictate success. Despite its early halt, Qwen3.7-Max delivers a highly competitive score and dominates the General tasks domain, demonstrating exceptional sample efficiency in synthesizing effective updates without exhaustive trial and error.

\paratitle{Cost analysis.}
Figure~\ref{fig:evolve-cost} illustrates the overall performance against the API inference cost incurred exclusively by the evolver model. The Pareto frontier reveals a steep logarithmic trade-off between financial investment and evolutionary capability. Achieving top-tier results demands substantial resources: GPT-5.6-Sol secures the highest score but dominates the cost spectrum by exceeding 500 USD per run. In stark contrast, models occupying the middle ground demonstrate exceptional cost-effectiveness. GLM-5.2 and Qwen3.7-Max establish the highly efficient knee of the curve, delivering formidable performance for under 40 USD. At the extreme budget end, DeepSeek-V4-Pro anchors the absolute baseline, successfully synthesizing functional harness improvements for less than a single dollar.


\subsection{Further Analysis: When Can Evolved Harnesses Surpass Human-Engineered Systems?}
\label{sec:analysis}

\begin{figure}[htbp]
    \centering
    \begin{minipage}[t]{0.54\linewidth}
        \vspace{0pt}
        \centering
        \includegraphics[width=\linewidth]{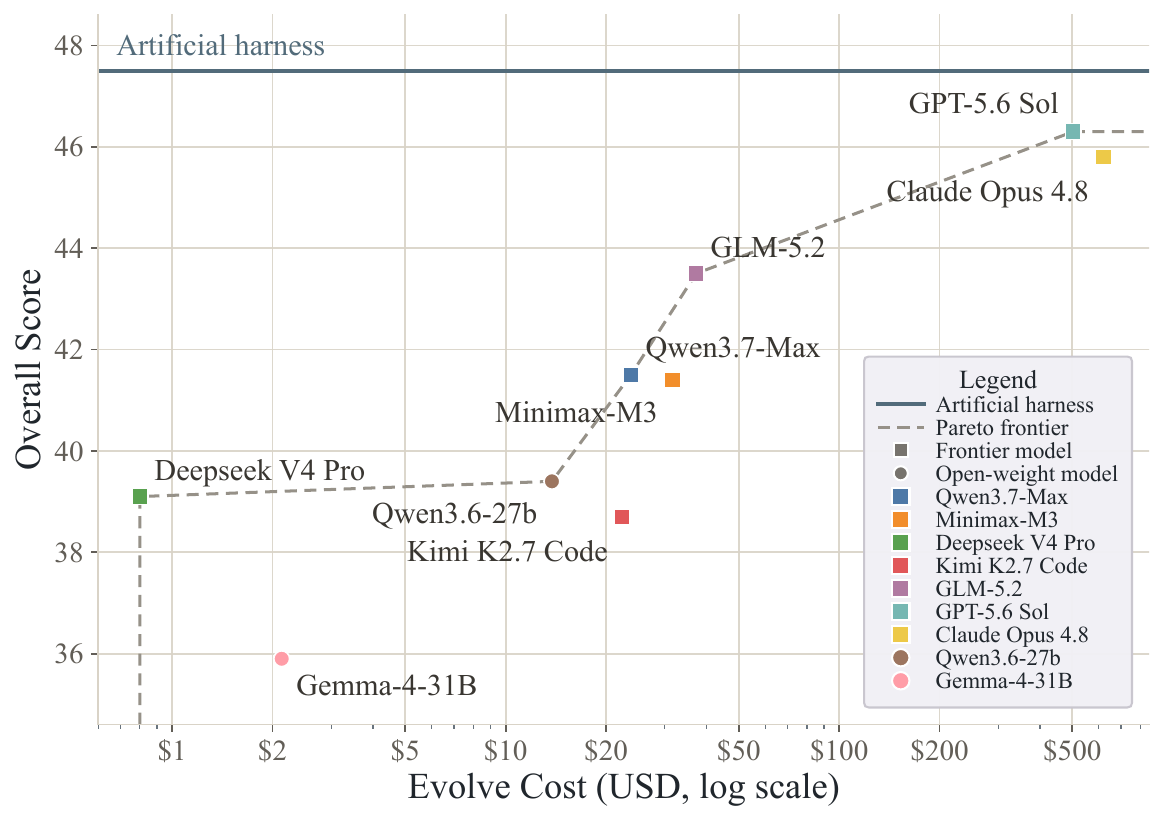}
        \caption{Cost(\$) vs. performance. The dashed line denotes the Pareto
        frontier; the horizontal line denotes the artificial harness.}
        \label{fig:evolve-cost}
    \end{minipage}
    \hfill
    \begin{minipage}[t]{0.43\linewidth}
        \vspace{0pt}
        \centering
        \includegraphics[width=\linewidth]{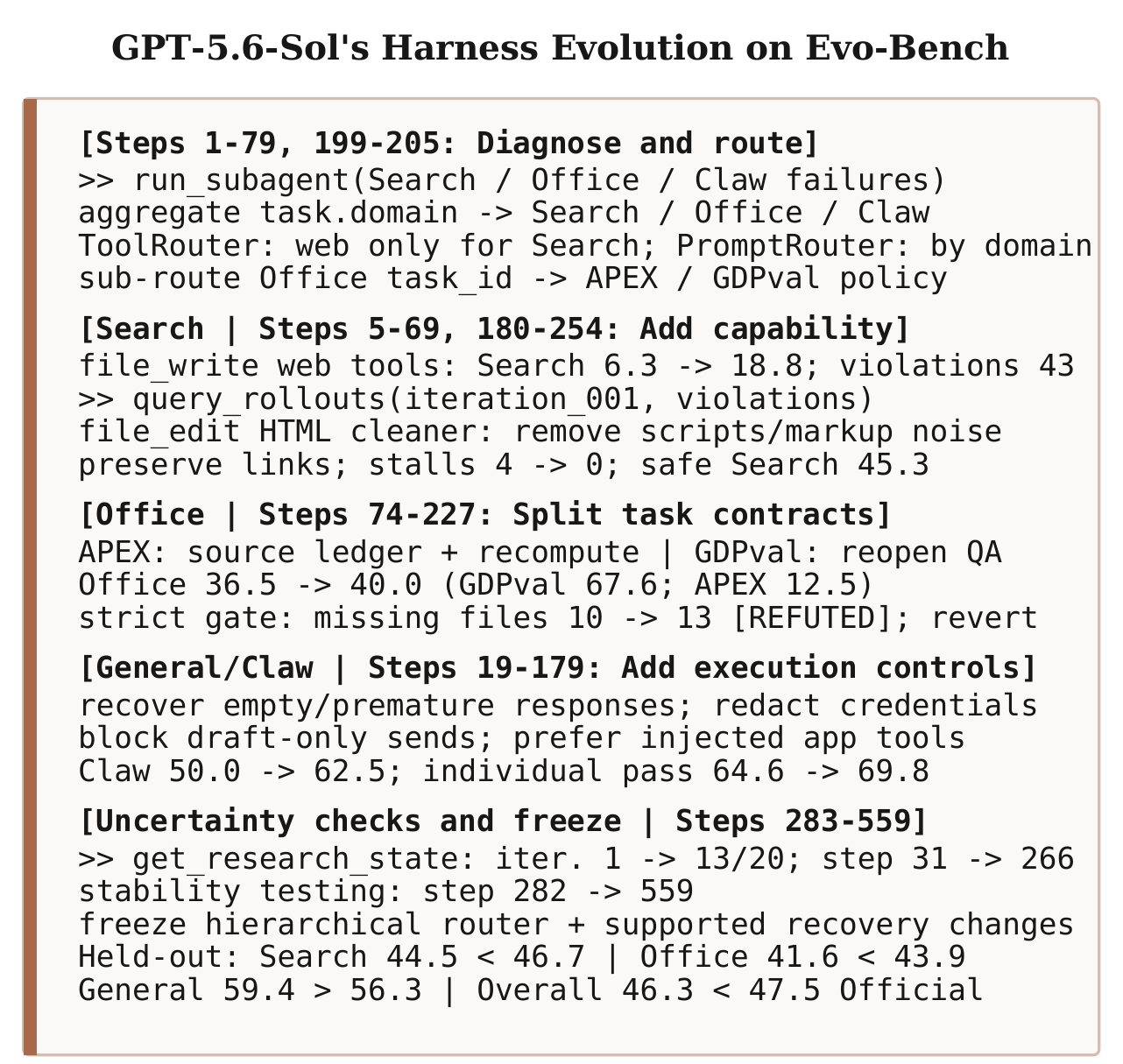}
        \caption{GPT-5.6-Sol harness-evolution case study.}
        \label{fig:gpt56-case-study}
    \end{minipage}
\end{figure}

\paratitle{Case study: how GPT-5.6-Sol evolves a harness.}
Figure~\ref{fig:gpt56-case-study} shows GPT-5.6-Sol acting as an autonomous systems engineer: instead of applying a uniform harness, it constructs a hierarchical router for domain-specific prompts and tools. It further demonstrates tool construction, failure diagnosis, and iterative correction. For Search, it implements web-search/fetch tools and a webpage cleaner that removes scripts, styles, and markup noise while preserving links. For Office, it separates APEX and GDPval: APEX tracks evidence by file, page, and sheet cell and recomputes key values, whereas GDPval requires artifact creation, recalculation, reopening, and rendered inspection. For General tasks, it adds recovery from empty or premature responses, credential redaction, and a guard against sending draft-only emails. Across evolvers, Opus-4.8 exhibits the closest systems-engineering profile to GPT-5.6-Sol, combining tool construction with domain-aware control, context management, failure recovery, and evidence-based reversion. GLM-5.2 demonstrates substantial diagnosis and tool engineering but relies more on domain-specific guidance and local controls, while Qwen3.7-Max and MiniMax M3 focus primarily on web tools, prompt specialization, and safety filters.

\paratitle{Challenges.}
Despite its achievements, GPT-5.6-Sol exposes three critical limitations. First, it reacts superficially to aggregate scores rather than distilling causal failure modes from extensive logs. Second, it resolves cross-domain interference via naive domain routing instead of discovering fundamentally robust, shared mechanisms. Third, it underutilizes the research budget, lacking the goal-oriented persistence of human engineers. Consequently, the evolved harness relies on localized modifications, leaving core components primitive: the planner acts passively, the context remains append-only, and the verifier stays entirely permissive. These behavioral patterns extend across other evolvers. Opus-4.8 exhibits the closest systems-engineering profile to GPT-5.6-Sol. In contrast, GLM-5.2 relies more heavily on local controls, while Qwen3.7-Max and MiniMax M3 focus primarily on basic web tools and prompt specialization.

\paratitle{Discussion.}
In summary, current frontier models demonstrate highly promising autonomous engineering capabilities, notably outperforming manual designs on General tasks. However, the overall lack of deep system-level optimization explains why they still trail the Artificial harness on Search with a score of 44.5 against 46.7, and on Office with 41.6 against 43.9. We believe that by strengthening the capacity to abstract causal failure patterns, execute large-scale architectural refactoring, and maximize budgets utilization, future language models hold substantial potential to continuously evolve harnesses that consistently surpass human experts.


\section{Ablation Studies}
\label{sec:ablation-studies}

To better understand the scalability and robustness of harness evolution, we conduct ablation studies examining two critical factors. 
First, we investigate how expanding the available resource budget impacts final performance. Second, we analyze whether the evolved harnesses successfully generalize across different underlying policy models. 

\subsection{Effect of Evolution Budget}

To evaluate the scaling behavior of harness evolution, we track the performance of Qwen3.7-Max and GLM-5.2 across three different budget constraints: 24 hours, 10 iterations with 500 steps; 36 hours, 15 iterations with 750 steps; and 48 hours, 20 iterations with 1000 steps. As illustrated in Figure~\ref{fig:budget-ablation}, both models demonstrate a clear positive scaling trend. Expanding the budget from 24 to 48 hours yields consistent, monotonic improvements in both the Overall Score and the Anytime Validation score.

Notably, GLM-5.2 exhibits a sharp performance trajectory, achieving a steep climb in its Anytime Validation metric during the initial expansion to 36 hours before plateauing. Conversely, Qwen3.7-Max demonstrates a steadier, linear growth across all measured constraints. This positive scaling trend indicates that despite the previously noted limitations in budget utilization, allocating greater computational resources for iterative exploration can still provide meaningful performance dividends, ultimately enabling the synthesis of more refined agent harnesses.

\subsection{Effect of the Policy Model}

Our main experiments fix DeepSeek-V4-Flash as the policy model. To verify whether the harness evolution capability is tied to a specific policy model, we swap it to Qwen3.6-35B-A3B and GLM-5.2, and re-evaluate the evolution process using Qwen3.7-Max and GLM-5.2 as evolvers. 

Table~\ref{tab:policy-model-ablation} demonstrates that harness evolution is highly robust to policy model changes. Regardless of whether the policy agent is based on Qwen, DeepSeek, or GLM, the evolved harnesses consistently achieve massive improvements over their respective CodeAct baselines. For instance, when utilizing Qwen3.6-35B-A3B as the policy model, the initial Overall score starts at a modest 13.9. Yet, both evolvers successfully optimize the harness to reach scores of 27.9 and 29.2. A similar trend emerges with the highly capable GLM-5.2 policy model, where the GLM-5.2 evolver lifts the baseline score from 38.0 to an impressive 48.4. These consistent cross-policy gains demonstrate that the evolvers are genuinely synthesizing generalizable reasoning and tool-use structures rather than merely overfitting to the idiosyncratic flaws of a single target model.

\definecolor{PolicyBlue}{HTML}{E3F5FB}

\begin{figure}[htbp]
\centering
\begin{minipage}[t]{0.47\linewidth}
    \vspace{0pt}
    \centering
    \includegraphics[width=\linewidth]{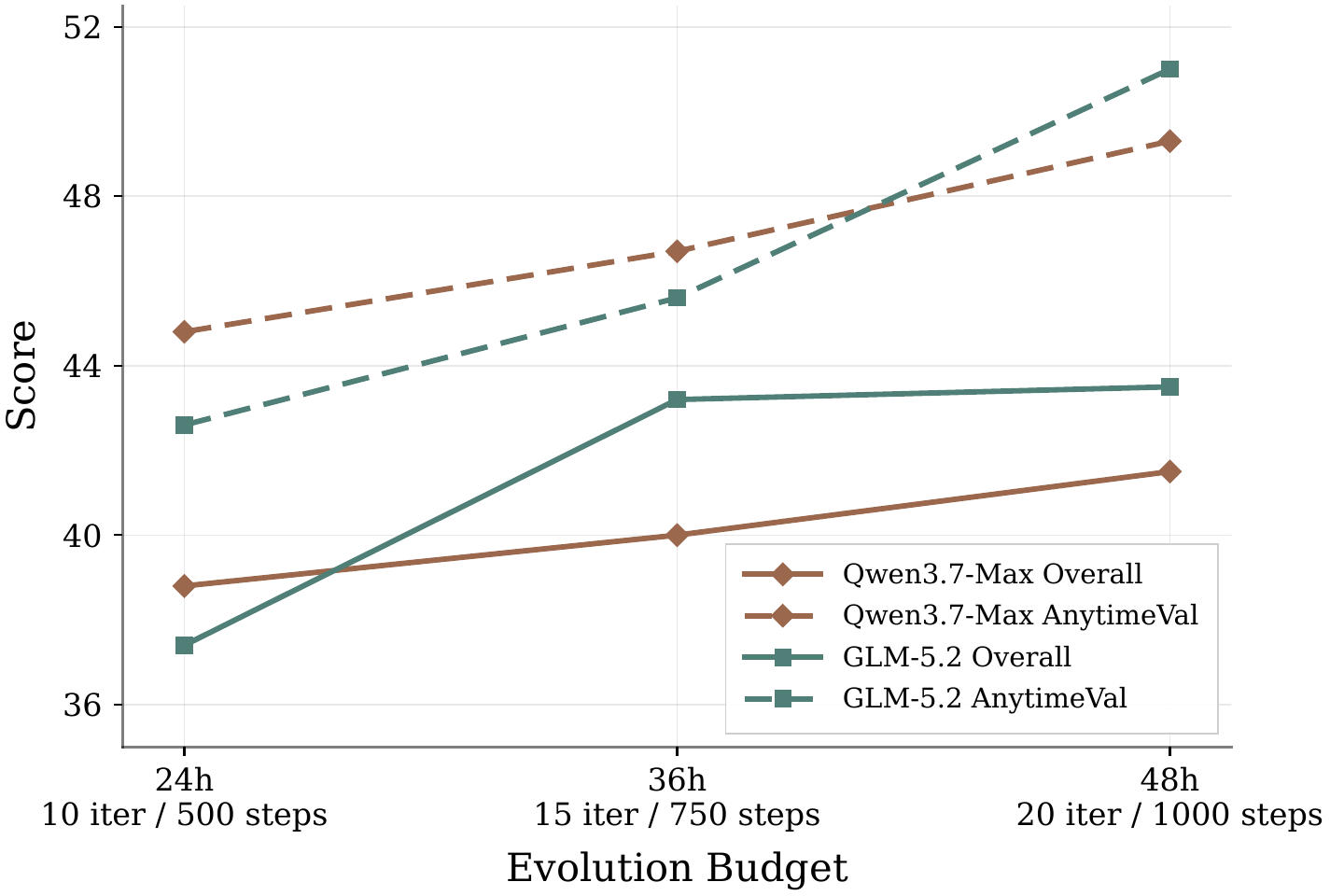}
    \caption{Effect of evolution budget.}
    \label{fig:budget-ablation}
\end{minipage}
\hfill
\begin{minipage}[t]{0.51\linewidth}
    \vspace{0pt}
    \centering
    \scriptsize
    \setlength{\tabcolsep}{1.8pt}
    \renewcommand{\arraystretch}{1.03}
    \begin{tabular}{@{}lrrrrr@{}}
    \toprule
    \textbf{Evolver}
    & \textbf{Search}
    & \textbf{Office}
    & \textbf{General}
    & \textbf{Overall}
    & \textbf{ATV} \\
    \midrule

    \rowcolor{PolicyBlue}
    \multicolumn{6}{c}{\textit{Policy: Qwen3.6-35B-A3B}} \\
    Baseline
    & 2.7 & 14.2 & 35.9 & 13.9 & -- \\
    Qwen3.7-Max
    & 12.5 & 33.0 & \textbf{48.4} & 27.9 & 29.2 \\
    GLM-5.2
    & \textbf{16.4} & \textbf{34.0} & 45.3
    & \textbf{29.2} & \textbf{30.8} \\
    \midrule

    \rowcolor{PolicyBlue}
    \multicolumn{6}{c}{\textit{Policy: DeepSeek-V4-Flash}} \\
    Baseline
    & 11.7 & 38.4 & 48.4 & 29.7 & -- \\
    Qwen3.7-Max
    & 36.3 & 37.8 & \textbf{59.4} & 41.5 & 49.3 \\
    GLM-5.2
    & \textbf{45.4} & \textbf{39.2} & 48.4
    & \textbf{43.5} & \textbf{51.0} \\
    \midrule

    \rowcolor{PolicyBlue}
    \multicolumn{6}{c}{\textit{Policy: GLM-5.2}} \\
    Baseline
    & 18.0 & 40.2 & 73.4 & 38.0 & -- \\
    Qwen3.7-Max
    & \textbf{38.3} & 45.1 & 46.9 & 42.7 & 46.7 \\
    GLM-5.2
    & 35.6 & \textbf{45.2} & \textbf{80.3}
    & \textbf{48.4} & \textbf{50.4} \\

    \bottomrule
    \end{tabular}
    \captionof{table}{Effect of the policy model on harness evolution. ATV
    denotes AnytimeVal; baseline is the CodeAct loop.}
    \label{tab:policy-model-ablation}
\end{minipage}
\end{figure}






\section{Conclusion}
In this work, we present Evo-Bench, a novel benchmark designed to evaluate language models' intrinsic capacity to autonomously improve agent harnesses. By establishing a rigorous framework anchored in controlled attribution, long-horizon iteration, and transfer alignment, Evo-Bench disentangles genuine harness-evolving capability from trivial prompt tuning and stochastic runtime noise. Our extensive evaluations across flagship LLMs reveal both the potential and current limitations of models operating as autonomous research engineers. We hope Evo-Bench serves as a foundational testbed to catalyze future research on self-evolving agent architectures and execution-grounded reasoning.
\section{Future Work}

Our goal is to maintain Evo-Bench as a living benchmark that continues to
provide an effective measure of harness evolution---an initial, practically
measurable form of AI self-evolution.   In the future, we will integrate coding tasks and challenging scientific research tasks into Evo-Bench, extend compatibility across diverse agent frameworks, and evaluate a wider range of models.

\bibliographystyle{iclr2027_conference}
\bibliography{paper}

\newpage
\appendix
\definecolor{appendixblue}{RGB}{43,86,166}
\newcommand{\appendixcontentssection}[3]{%
  \par\noindent
  \hyperref[#3]{{\color{appendixblue}\large\bfseries
   \makebox[2.7em][l]{#1}#2}}%
  \nobreak\hfill
  \hyperref[#3]{{\large\bfseries\pageref*{#3}}}\par
  \vspace{0.55ex}%
}
\newcommand{\appendixcontentssubsection}[3]{%
  \par\noindent\hspace*{2.7em}%
  \hyperref[#3]{{\color{appendixblue}\makebox[3.6em][l]{#1}#2}}%
  \nobreak\leaders\hbox{\kern0.28em.\kern0.28em}\hfill\nobreak
  \hyperref[#3]{\makebox[2.2em][r]{\pageref*{#3}}}\par
  \vspace{0.22ex}%
}

\pdfbookmark[0]{Appendix}{appendix-contents}
\begingroup
\hypersetup{hidelinks}
\begin{center}
  {\color{appendixblue}\fontsize{20}{28}\selectfont\bfseries
   Appendix}
\end{center}
\vspace{2.2em}

\appendixcontentssection{A}{Benchmark Details}{app:benchmark-details}
\appendixcontentssubsection{A.1}{Details of the Evolve Harness}{app:evolve-harness}
\appendixcontentssubsection{A.2}{Details of the Policy Harness}{app:policy-harness}
\appendixcontentssubsection{A.3}{Benchmark Difficulty and Harness Sensitivity}{app:benchmark-item-map}

\vspace{1.4ex}
\appendixcontentssection{B}{Details of Benchmark Construction}{app:stage1-details}
\appendixcontentssubsection{B.1}{Selection of Auxiliary Tasks}{app:auxiliary-task-selection}
\appendixcontentssubsection{B.2}{Selection of Auxiliary Harnesses}{app:auxiliary-harness-selection}

\vspace{1.4ex}
\appendixcontentssection{C}{Experiment Details}{app:experiment-details}
\appendixcontentssubsection{C.1}{Model Configurations and Cost Accounting}{app:model-configurations}
\appendixcontentssubsection{C.2}{Integrity and Reward-Hacking Analysis}{app:integrity-analysis}

\vspace{1.4ex}
\appendixcontentssection{D}{Failure Mode Analysis of Evolution Trajectories}{app:failure-modes}
\appendixcontentssubsection{D.1}{Qwen3.6-27B: Misattributing Regressions to Noise}{app:qwen-failure-mode}
\appendixcontentssubsection{D.2}{DeepSeek-V4-Pro: Premature Plateau and Procedural Evaluations}{app:deepseek-failure-mode}
\appendixcontentssubsection{D.3}{Kimi-K2.7-Code: Careful Rollback but Local Search Saturation}{app:kimi-failure-mode}
\appendixcontentssubsection{D.4}{Cross-Model Failure Patterns}{app:cross-model-failures}

\vspace{1.4ex}
\appendixcontentssection{E}{Prompt Design}{app:prompt-design}
\appendixcontentssubsection{E.1}{Evolver Prompt}{app:evolver-prompt}
\appendixcontentssubsection{E.2}{Seed Policy Prompt}{app:seed-policy-prompt}
\appendixcontentssubsection{E.3}{Judge Prompts}{app:judge-prompts}
\endgroup

\newpage

\section{Benchmark Details}
\label{app:benchmark-details}

This section supplements the benchmark description with implementation and diagnostic details.  It describes the fixed evolve harness, the seed policy harness, and the task-level difficulty and harness-sensitivity characteristics of the constructed benchmark suites.

\subsection{Details of the Evolve Harness}
\label{app:evolve-harness}

The fixed evolve harness $\mathcal{H}_{\mathrm{evo}}$ lets every evolver inspect,
revise, and evaluate a policy harness under the same prompt, tools, state
management, and resource accounting.  Table~\ref{tab:evolve-harness-components}
summarizes its components.

\begin{table}[htbp]
\centering
\small
\caption{Components of the evolve harness.}
\label{tab:evolve-harness-components}
\begin{tabularx}{\textwidth}{p{0.21\textwidth} X}
\toprule
\textbf{Component} & \textbf{Function} \\
\midrule
Evolution orchestrator & Defines the optimization objective, hypothesis-driven
workflow, editable surfaces, integrity rules, and termination protocol. \\
Engineering tools & Provide file inspection and editing, code search, shell
execution, web access, and read-only delegated exploration. \\
Research skills & Support rollout slicing, behavioral comparison, multi-eval
analysis, experiment logging, insight retention, and architecture-level
reflection. \\
Evaluation interface & Evaluates immutable harness snapshots asynchronously and
returns scores, per-domain results, task outcomes, and failure diagnostics. \\
\makecell[l]{Context and state\\management} & Tracks budgets, evaluations, best revisions, and a persistent
experiment ledger, while compacting old interaction history for long runs. \\
\makecell[l]{Sandbox and snapshot\\management} & Separates evolver evidence from policy inputs, records
revisions and trajectories, and freezes one harness for held-out evaluation. \\
\bottomrule
\end{tabularx}
\end{table}

\paratitle{Evolution orchestrator.}
The orchestrator separates the editable policy harness from a scratch
workbench and promotes a diagnose--hypothesize--edit--evaluate cycle over all
harness surfaces.  A live state view reports evaluation, step, time, and token
budgets, recent scores, and experiment history; limits are enforced by the
harness.

\paratitle{Engineering tools.}
File, search, shell, and web tools support implementation and external research.
A read-only subagent can inspect code, validation artifacts, and web sources in
an independent context, but cannot edit or evaluate the harness.

\paratitle{Research skills.}
Six progressively disclosed skills provide rollout filtering
(\texttt{query\_rollouts}), task-paired comparison (\texttt{diff\_task}), and
multi-evaluation analysis (\texttt{analyze\_evals}).  The ledger skills record
hypotheses (\texttt{log\_experiment}), close them from evidence
(\texttt{record\_insight}), and trigger broader redesign after repeated local
failures (\texttt{architecture\_checkpoint}).  All are optional views over raw,
evolver-visible artifacts.

\paratitle{Evaluation interface.}
\texttt{run\_train\_eval} consumes one iteration and asynchronously evaluates an
immutable snapshot; at most one evaluation is active.  \texttt{await\_eval}
returns the headline metric, changes from previous and best revisions,
per-domain results, worst tasks, and failure classes.  Full results, feedback,
logs, and trajectories remain available.

\paratitle{Context and state management.}
The harness records interactions, tool calls, evaluations, resource use, and
snapshots; the ledger persists resolved hypotheses.  Near the context limit,
older messages are summarized while recent turns and all disk artifacts remain
available.

\paratitle{Sandbox and snapshot management.}
The evolver may read validation evidence but write only the policy harness and
workbench.  Policy rollouts receive a read-only snapshot, public task, and
isolated workspace, never answers, scorers, evolver files, or held-out data.
Integrity checks scan source and trajectories; held-out evaluation begins only
after the selected revision is frozen.

\subsection{Details of the Policy Harness}
\label{app:policy-harness}

\begin{figure}[htbp]
    \centering
    \begin{tikzpicture}[
        node distance=1.35cm and 1.55cm,
        basebox/.style={rounded corners=2pt, minimum height=0.72cm,
                        minimum width=2.45cm, align=center, font=\small},
        neutralbox/.style={basebox, draw=black!55, fill=black!4},
        modelbox/.style={basebox, draw=blue!65, fill=blue!7},
        shellbox/.style={basebox, draw=teal!65, fill=teal!7},
        flow/.style={-{Latex[length=2mm]}, line width=0.8pt, draw=black!65},
        label/.style={font=\scriptsize, text=black!70, fill=white,
                      inner sep=1.5pt}
    ]
        \node[neutralbox] (task) {Public task\\and workspace};
        \node[modelbox, right=of task] (model) {Fixed policy model};
        \node[shellbox, right=of model] (shell) {\texttt{run\_shell\_command}};
        \node[neutralbox, below=0.95cm of model] (answer) {Final answer\\or artifact};

        \draw[flow] (task) -- node[label, above]{instruction} (model);
        \draw[flow] ([yshift=4pt]model.east) --
            node[label, above]{command} ([yshift=4pt]shell.west);
        \draw[flow] ([yshift=-4pt]shell.west) --
            node[label, below]{observation} ([yshift=-4pt]model.east);
        \draw[flow] (model) -- node[label, right]{\texttt{finish}} (answer);
    \end{tikzpicture}
    \caption{The minimal CodeAct policy harness. Shell observations are appended
    to the model context until the task is completed or a resource limit is
    reached.}
    \label{fig:seed-policy-harness}
\end{figure}
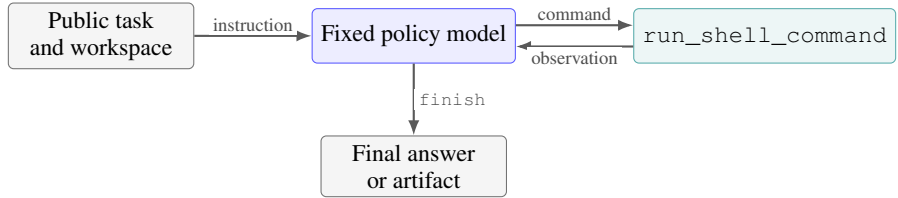

All runs start from the CodeAct seed policy harness
$H_0$~\cite{wang2024executable} in Figure~\ref{fig:seed-policy-harness}.  The
fixed policy model alternates shell commands and append-only observations until
\texttt{finish}, a tool-used answer, or a resource limit.  Domain tools,
planning, memory, and verification are left for evolution.

\subsection{Benchmark Difficulty and Harness Sensitivity}
\label{app:benchmark-item-map}

\begin{figure*}[htbp]
    \centering
    \includegraphics[width=\textwidth]{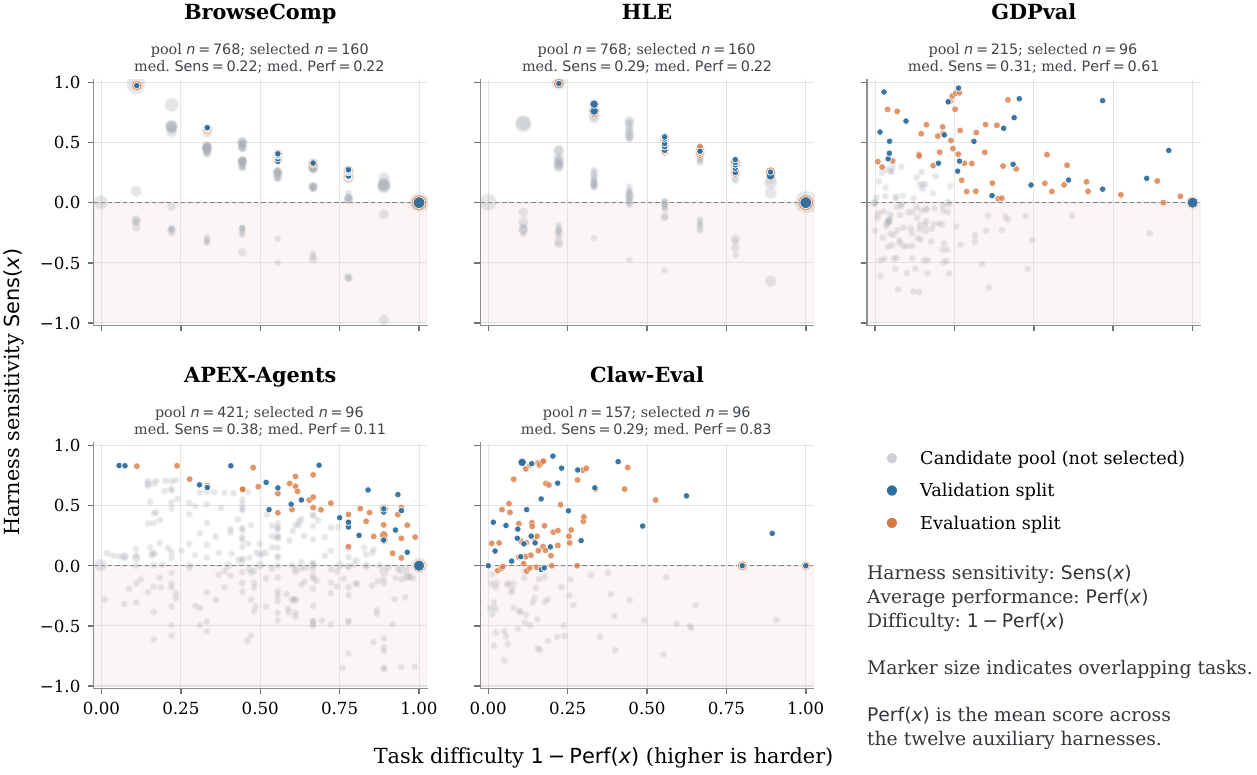}
    \caption{Difficulty and harness sensitivity of candidate and selected benchmark
    tasks.  Gray, blue, and orange denote unselected, validation, and evaluation
    tasks; marker size denotes coordinate multiplicity.  Annotations give pool
    and selected sizes, and shading marks negative sensitivity.}
    \label{fig:alpha-beta-item-map}
\end{figure*}

Figure~\ref{fig:alpha-beta-item-map} plots harness sensitivity
$\mathrm{Sens}(x)$ and difficulty $1-\mathrm{Perf}(x)$, where
$\mathrm{Perf}(x)$ is mean performance across the twelve auxiliary harnesses.
Selection favors non-negative sensitivity while preserving difficulty coverage,
and the two splits remain aligned.  BrowseComp, HLE, and APEX-Agents skew hard;
GDPval is broader and Claw-Eval easier.  Eight slightly negative Claw-Eval tasks
are quota backfills from sparse strata.

\section{Details of Benchmark Construction}
\label{app:stage1-details}
Section~\ref{sec:calibration-tasks} summarizes the two steps of Stage~1: we
first select a 320-task auxiliary suite and then use it to generate and select a
representative set of auxiliary harnesses.  This section provides the full
selection details for both components.  Section~\ref{app:auxiliary-task-selection}
describes the selection of auxiliary tasks, and
Section~\ref{app:auxiliary-harness-selection} describes the selection of
auxiliary harnesses.

\subsection{Selection of Auxiliary Tasks}
\label{app:auxiliary-task-selection}

The first step of Stage~1 selects the auxiliary tasks used to induce harness
diversity.  We
collect Search from MiroRL~\cite{2025mirorl}, RedSearcher~\cite{redsearcher2026},
and internal corpora; Office from anonymized enterprise workflows; and General
from Auto-ClawEval~\cite{li2026clawenvkitautomaticenvironmentgeneration}.  After
validity, rollout-coverage, deduplication, and cross-suite leakage checks, 11,322
candidates remain: 2,988 Search, 7,564 Office, and 770 General.

\paratitle{Rollout-based characterization.}
Using DeepSeek-V4-Flash with the domain-native MiroFlow~\cite{miromind2025mirothinker},
Stirrup~\cite{artificialanalysis2026stirrup}, and Claw-Eval~\cite{claweval}
workflows, we estimate mean score $s(x)$ and turns $\ell(x)$.  We require two
judged Search rollouts and three valid Office/General rollouts.  This process is
independent of the CodeAct seed.

\begin{figure}[htbp]
    \centering
    \includegraphics[width=1.0\linewidth]{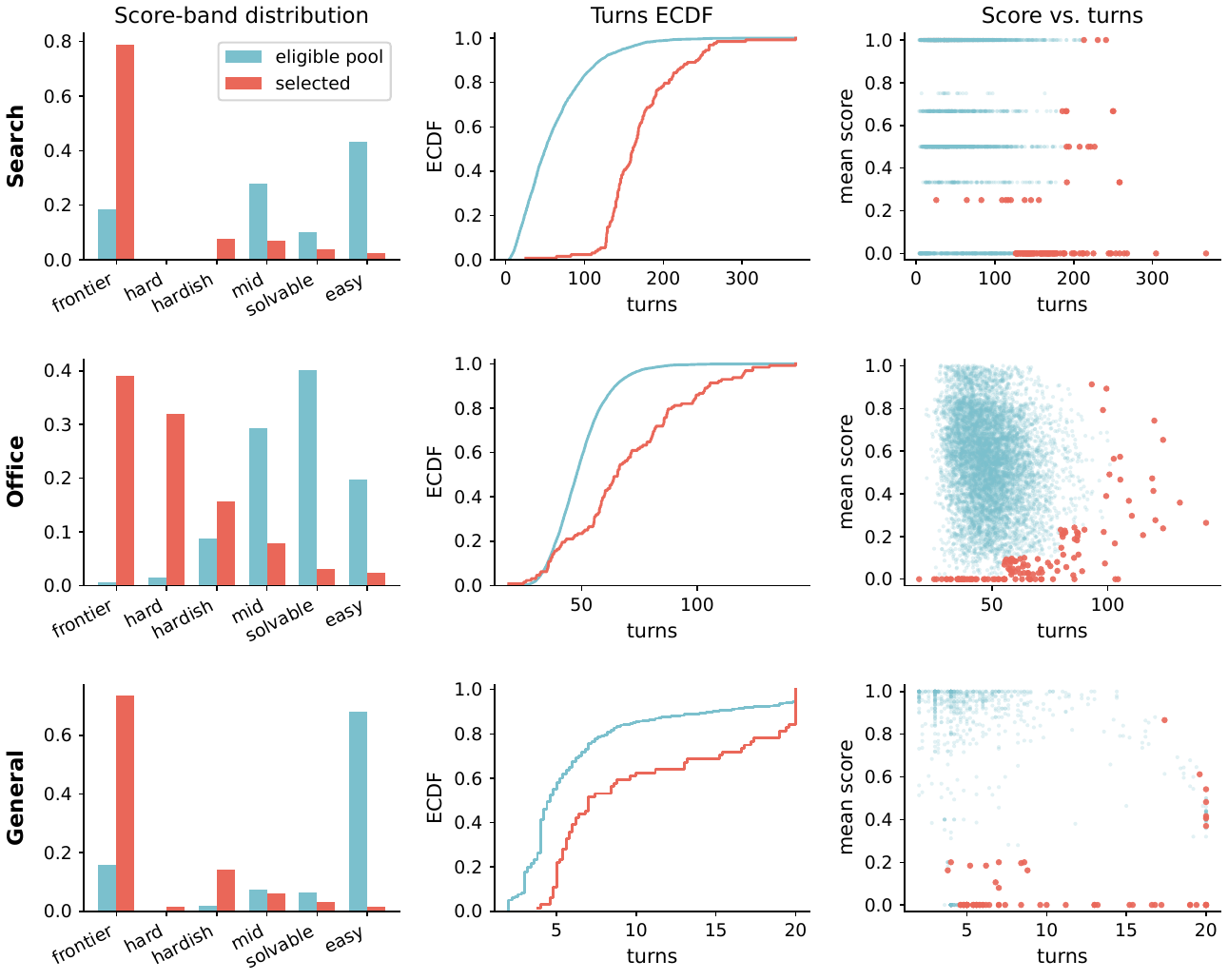}
    \caption{Auxiliary-task selection by headroom and interaction horizon.
    Rows are Search, Office, and General; columns show score bands, turn CDFs,
    and score--turn distributions.  Blue is the candidate pool and red the
    selected 320-task suite.}
    \label{fig:auxiliary-task-selection}
\end{figure}

\paratitle{Headroom--horizon selection.}
Within each domain, tasks enter six score bands:
\emph{frontier} ($s=0$), \emph{hard} ($0<s\leq0.10$), \emph{hardish}
($0.10<s\leq0.25$), \emph{mid} ($0.25<s\leq0.50$), \emph{solvable}
($0.50<s\leq0.75$), and \emph{easy} ($s>0.75$), with quota weights
$(0.58,0.24,0.11,0.045,0.02,0.005)$.  Deficits are redistributed, and candidates
within each band are ranked by decreasing $\ell(x)$ with seed 20260716.  This
selects 128 Search, 128 Office, and 64 General tasks, favoring headroom and long
interaction horizons (Figure~\ref{fig:auxiliary-task-selection}).

\subsection{Selection of Auxiliary Harnesses}
\label{app:auxiliary-harness-selection}
After selecting the auxiliary tasks, the second step of Stage~1 uses them to
generate a diverse pool of harness revisions and select a representative
auxiliary harness set.
Claude Opus-4.8, Claude Sonnet-5, GLM-5.2, and GPT-5.6-Sol independently evolve
the same seed CodeAct harness on the 320 auxiliary tasks with
DeepSeek-V4-Flash as the fixed policy model and the main budget (20 iterations,
1,000 steps, 48 hours).  Holding the starting harness, policy model, evolution
protocol, and resource budget constant isolates variation induced by the four
evolvers.  Together, the four evolutions produce 73 evaluated harnesses for the
subsequent selection procedure.

\paratitle{Snapshot deduplication and representation.}
Because successive iterations can produce code-identical harnesses, we first
deduplicate the candidate pool.  Content hashing reduces the 73 evaluated
harnesses to 65 unique revisions; the common seed harness is excluded.  To
compare the remaining revisions on a common basis, we represent each one by its
exercised and coded capabilities, observed tool use, program structure, and
auxiliary performance.  The capability representation covers tooling, context,
control flow, verification, memory, recovery, multi-agent execution, and
routing.

\paratitle{Representative-anchor selection.}
We then select a balanced subset that preserves the progression within each
evolution while covering distinct harness designs.  A deterministic,
final-snapshot-constrained $k$-medoids procedure selects three harnesses per
evolution.  It retains the final revision, preserves capability levels seen in
at least two revisions, minimizes representation error, and breaks ties by
pairwise diversity.  Selecting the same number of harnesses from each evolution
prevents evolutions with more evaluated revisions from dominating the selected
set.  The four evolutions therefore yield 12 harnesses,
$\mathcal{H}_{\mathrm{aux}}$, completing the auxiliary-harness selection step
of Stage~1.  As described in Section~\ref{sec:harness-response-prior},
candidate-task evaluations under this set are subsequently used in Stage~2 to
compute $\mathrm{Sens}(x)$ and $\mathrm{Perf}(x)$. 

\section{Experiment Details}
\label{app:experiment-details}
This section supplements the experimental setup and reporting protocol.  It documents model configurations and cost accounting, followed by the integrity controls and trajectory audits used to detect and correct reward-hacking behavior.

\subsection{Model Configurations and Cost Accounting}
\label{app:model-configurations}

We evaluate seven frontier and two open-weight evolvers
(Table~\ref{tab:model-configurations}) using thinking mode, the largest supported
context, and the highest exposed reasoning effort.  Requests allow 65,536
output tokens and time out after 600 seconds; default
temperature is $1.0$ unless endpoint-controlled.  Policy and judge are fixed.

\begin{table}[htbp]
\centering
\small
\caption{Model configurations used in the main experiments.  ``Enabled''
denotes a model-specific thinking switch rather than a scalar reasoning-effort
setting.  ``Default'' indicates endpoint-controlled temperature.}
\label{tab:model-configurations}
\begin{tabular*}{\textwidth}{@{\extracolsep{\fill}}lllcc@{}}
\toprule
\textbf{Role} & \textbf{Model} & \textbf{Reasoning} & \textbf{Temp.} & \textbf{Context} \\
\midrule
Evolver & Qwen3.7-Max       & \texttt{xhigh} & 1.0 & 1M \\
Evolver & MiniMax M3        & \texttt{max}   & 1.0 & 1M \\
Evolver & DeepSeek-V4-Pro   & \texttt{max}   & 1.0 & 1M \\
Evolver & Kimi-K2.7-Code    & Enabled        & Default & 256K \\
Evolver & GLM-5.2           & \texttt{max}   & 1.0 & 1M \\
Evolver & GPT-5.6-Sol       & \texttt{max}   & 1.0 & 384K \\
Evolver & Claude Opus-4.8   & \texttt{max}   & 1.0 & 1M \\
Evolver & Qwen3.6-27B       & \texttt{xhigh} & 1.0 & 256K \\
Evolver & Gemma 4 31B       & Enabled        & 1.0 & 256K \\
\midrule
Policy (fixed) & DeepSeek-V4-Flash & \texttt{max} & 1.0 & 256K \\
Judge (fixed)  & Qwen3.7-Plus      & Standard & 0.0 & 1M \\
\bottomrule
\end{tabular*}
\end{table}
\paratitle{Fixed policy and judge.}
DeepSeek-V4-Flash uses maximum reasoning, 256K context, the same output cap, and
a 600-second timeout.  Qwen3.7-Plus grades model-judged tasks at temperature
$0.0$, 1M context, the same cap, and a 1,200-second timeout.  Claude Opus-4.8
uses Anthropic messages; other models use OpenAI-compatible interfaces.

\paratitle{Cost accounting.}
From per-call usage, model $m$ costs
\begin{equation*}
C_m=\frac{1}{10^6}\sum_r
\left(n^{\mathrm{in}}_r p^{\mathrm{in}}_m
+n^{\mathrm{cr}}_r p^{\mathrm{cr}}_m
+n^{\mathrm{cw}}_r p^{\mathrm{cw}}_m
+n^{\mathrm{out}}_r p^{\mathrm{out}}_m\right),
\end{equation*}
where $n$ counts uncached input, cache-read, cache-write, and output tokens, and
$p$ is the corresponding public non-batch list price per million tokens on July
10, 2026.  Request length selects tiered prices.  The cost--performance analysis
includes main-loop, compaction, and subagent calls but excludes policy and judge
costs; full decompositions are retained.  Missing cache counters are charged as
uncached input.

\subsection{Integrity and Reward-Hacking Analysis}
\label{app:integrity-analysis}

Harness evolution deliberately gives the evolver broad autonomy within an independent sandbox.  It can inspect rollout trajectories, task provenance, expected answers, and validation scores; use shell and web tools; and modify the policy harness.  This access is necessary for diagnosing failures and developing general harness improvements, but it also creates substantial opportunities for reward hacking.  During our experiments, we focused on three main forms of hacking behavior and introduced a corresponding control for each one.

\paratitle{(1) Leakage across the evolver--policy boundary.}
An evolver could misuse validation access by transferring protected information into policy execution or by attempting to interfere with the evaluation protocol.  We control this risk through strict sandbox separation.  The evolver may read validation artifacts but can write only to its policy harness and workbench.  Each policy rollout instead receives an immutable harness snapshot, a public task view with answers and scorer fields removed, and a fresh workspace; validation artifacts, evolver notes, sibling workspaces, and held-out data are not mounted.  Scorers, aggregation logic, suite definitions, and iteration accounting remain in the fixed host-side framework.  Every formal validation evaluation consumes one iteration, and held-out evaluation starts only after the selected harness is frozen and returns no evidence to the evolver.

\paratitle{(2) Retrieval of benchmark answers.}
A model with web access may search for benchmark-specific answers, answer keys, or evaluation artifacts instead of solving a task from permitted evidence. Simply disabling web access is not appropriate because retrieval is intrinsic to Search tasks.  We therefore scan content returned by web and shell tools for known patterns associated with benchmark datasets, answer keys, reference artifacts, and evaluation dumps.  The check applies to information delivered to the policy rather than to attempted queries: an unsuccessful request is not penalized, and a request blocked before protected content reaches the policy is not treated as a violation.  For a confirmed match, we record the source and tool event, mark the trial as a policy violation, and set all native score fields to zero before computing domain scores and Overall Score.

\paratitle{(3) Evasion of the retrieval safeguard.}
A model may also attempt to bypass the preceding scanner by paraphrasing, encoding, obfuscating, or concealing retrieved information.  Because pattern matching alone cannot reliably detect such behavior, we apply the same Codex-based semantic audit to the complete evolver and policy trajectories of every evolution run.  The audit checks whether protected information reached policy execution, whether retrieval targeted benchmark-specific artifacts, and whether behavior attempted to conceal detector-visible evidence.  Among all evaluated models, only MiniMax M3 exhibited this detector-evasion behavior.  The audit identified the affected trials; we set their scores to zero, recomputed the aggregates, and report the corrected MiniMax M3 result.  Tool calls, observations, harness snapshots, and violation records are retained for inspection.

\section{Failure Mode Analysis of Evolution Trajectories}
\label{app:failure-modes}

We present the complete evolution process of three lower-scoring models and analyze the failure modes that limited their final performance.  The analysis draws on transcripts, experiment ledgers, validation digests, and harness snapshots.  Scores are validation Overall Scores on the same 0--100 scale used in the main experiments; I$i$ denotes iteration $i$, and evaluation-suite results are excluded.

\begin{figure}[htbp]
    \centering
    \includegraphics[width=\linewidth]{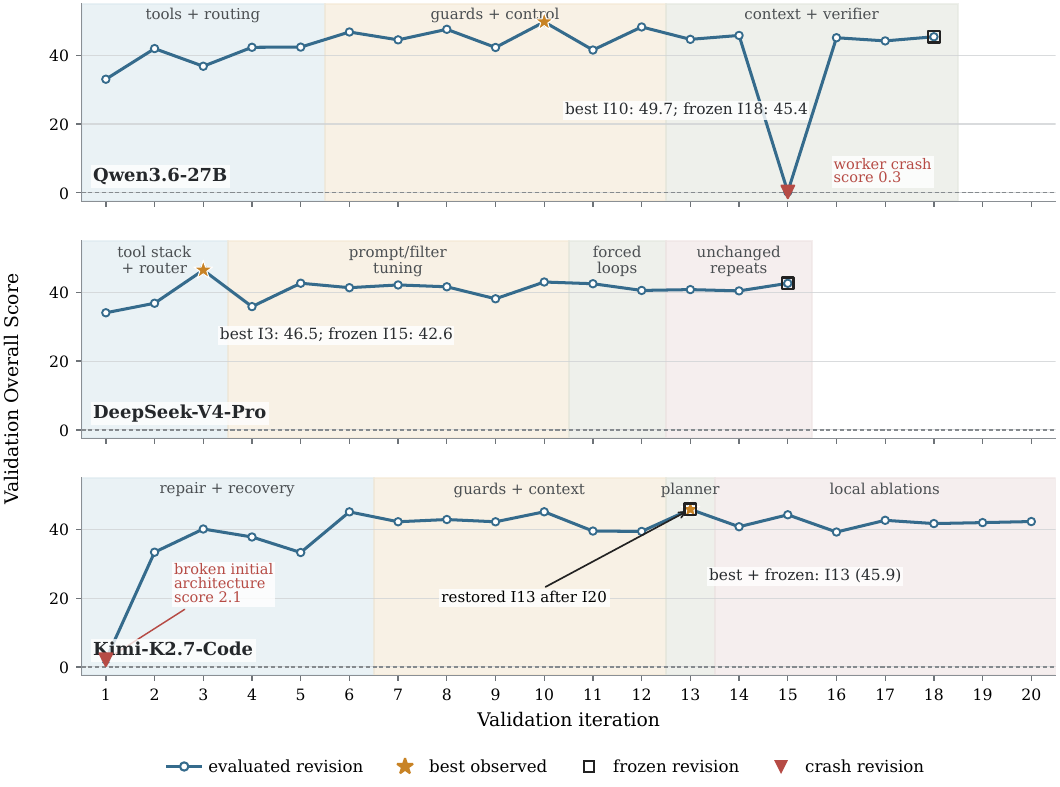}
    \caption{Validation Overall Score trajectories.  Bands show intervention
    surfaces; stars and squares mark best and frozen revisions.  Catastrophic
    revisions are annotated.  Kimi restored I13 after I20.}
    \label{fig:failure-mode-trajectories}
\end{figure}

\begin{table}[htbp]
\centering
\caption{Research-budget use and revision selection.  ``Rollout accesses''
counts raw-artifact accesses; Kimi restored the byte-identical I13 snapshot
after I20.}
\small
\begin{tabular*}{\linewidth}{@{\extracolsep{\fill}}lcccc@{}}
\toprule
Evolver & Iterations & Steps & Rollout accesses & Best $\rightarrow$ frozen \\
\midrule
Qwen3.6-27B      & 18/20 & 349/1000 & 16 & 49.7 (I10) $\rightarrow$ 45.4 (I18) \\
DeepSeek-V4-Pro  & 15/20 & 202/1000 & 4  & 46.5 (I3) $\rightarrow$ 42.6 (I15) \\
Kimi-K2.7-Code   & 20/20 & 549/1000 & 52 & 45.9 (I13) $\rightarrow$ 45.9 (I13) \\
\bottomrule
\end{tabular*}
\label{tab:failure-mode-budget}
\end{table}
\subsection{Qwen3.6-27B: Misattributing Regressions to Noise}
\label{app:qwen-failure-mode}

Qwen rapidly added retrieval, file/Python tools, routing, integrity filters,
verification, and recovery, rising from 33.0 to 49.7 at I10.  Later it bundled
context and verifier changes; a malformed I15 verifier crashed 159/160 tasks
and reduced the score to 0.3, a defect missed by local preflight.  Repairs
recovered only to 45.4 at I18.

Despite a 2.2-point Overall Score range across byte-identical I8/I10/I12
revisions, Qwen attributed the 4.3-point I10--I18 gap largely to noise without
replicating or restoring I10/I12.
It froze I18 with two iterations and 651 steps unused.  The failure combines
bundled edits, missing smoke checks, and weak best-snapshot management.

\subsection{DeepSeek-V4-Pro: Premature Plateau and Procedural Evaluations}
\label{app:deepseek-failure-mode}

DeepSeek reached 46.5 at I3 after adding search, fetch, file/Python tools, URL
filters, and domain guidance.  It then tuned prompts, filters, and completion
rules while accessing raw rollouts only four times; I4--I10 ranged from 35.9 to
43.0.  Requiring three searches at I11 reduced the score to 42.5 and tripled
runtime; I12 reflection reached 40.6.

After an attempted early stop, I13--I15 scored 40.8, 40.4, and 42.6, with the
last two evaluating unchanged code.  DeepSeek labeled I3 an outlier without
reevaluating it and froze I15 with five iterations and 798 steps unused.  This
is aggregate hill climbing without causal localization, followed by procedural
budget use and premature plateau declaration.

\subsection{Kimi-K2.7-Code: Careful Rollback but Local Search Saturation}
\label{app:kimi-failure-mode}

Kimi recovered from an invalid I1 architecture (score 2.1) to 45.1 at I6,
tested and reverted regressive context and finish controls, then reached 45.9
at I13 with a step-budget planner.  Seven further tests of planning, APEX
instructions, thinking, checklists, step limits, and verification did not
improve it.  Kimi logged these results, restored byte-identical I13, and froze
the best revision.

Its limitation was search efficiency: despite exhausting all 20 evaluation
iterations and accessing 52 rollouts, it used only 549/1,000 steps.  Post-I13
work remained local prompt, gate, context, and threshold variants rather than
deeper planning or verification redesign.  Rollback discipline prevented
regression but not architectural saturation.

\subsection{Cross-Model Failure Patterns}
\label{app:cross-model-failures}

All three add obvious capabilities quickly but later favor local prompts,
filters, thresholds, and gates over causal, task-paired redesign.  Qwen and
DeepSeek stop early; Kimi exhausts evaluations on low-yield search.  Only Kimi
restores its best snapshot.  Stronger evolution therefore needs paired failure
analysis, one falsifiable mechanism per iteration, cheap preflight checks,
automatic best-revision recovery, and an architectural-reset trigger after
repeated local failures.

\section{Prompt Design}
\label{app:prompt-design}

This section summarizes the prompts used throughout the Evo-Bench evaluation process.  They comprise a fixed evolver prompt, an editable policy prompt, and fixed evaluation-side judge prompts.  Evolvers may modify only the policy layer; policies never receive references, rubrics, or judge instructions.

\subsection{Evolver Prompt}
\label{app:evolver-prompt}

All nine evolvers receive the same template below, apart from serialization.
Bracketed fields supply domains, metric, baseline, and budget; only repeated
motivational text and expanded schemas/catalogs are omitted.

\begin{tcolorbox}[
    breakable,
    colback=black!2,
    colframe=black!35,
    boxrule=0.5pt,
    arc=1mm,
    left=1.5mm,
    right=1.5mm,
    top=1mm,
    bottom=1mm,
    title={Fixed evolver prompt template},
    fonttitle=\bfseries\small]
\footnotesize
\textbf{Identity and objective.}
You are a harness research engineer operating in an isolated sandbox.  Improve
the editable policy harness to maximize the held-out headline metric.  Use
validation only, prefer general mechanisms, and diagnose trajectories before
structural changes; treat regressions as evidence.

\textbf{Boundaries and integrity.}
The policy harness at \texttt{[POLICY]} is the product that will be frozen and
scored; \texttt{[WORKBENCH]} is unscored scratch space.  Validation answers,
feedback, and rollouts are diagnostic evidence, not policy resources.  Do not
modify evaluation, transfer protected evidence, exploit artifacts, or conceal
policy actions.

\textbf{Tools.}
Use engineering tools and read-only subagents for scoped research;
\texttt{get\_research\_state} for budgets;
\texttt{run\_train\_eval}/\texttt{await\_eval} for formal evaluation; and
research skills for rollout analysis and the experiment ledger.  Call
\texttt{complete\_evolution} only after restoring the selected revision.

\textbf{Experimental workflow.}
Treat the clean seed as iteration 0.  Repeatedly inspect failures, state a
falsifiable mechanism, edit and locally test, evaluate an immutable snapshot,
and retain or revert from evidence.  Any harness surface may be redesigned.
Analyze while evaluations run and restore the best revision before freezing.

\textbf{Opening run specification.}
The run contains \texttt{[DOMAIN DESCRIPTIONS]} and optimizes
\texttt{[HEADLINE METRIC]}.  The clean baseline is
\texttt{[BASELINE DIGEST]}.  Redesign the minimal seed within the evaluation,
step, and time budgets; evaluation-suite tasks are unavailable.

\textbf{Research state.}
\texttt{iteration [USED]/[MAX]; step [USED]/[MAX]; [TIME] remaining;
tokens [COUNT]; current/best score [VALUES]; active evaluation [STATE].}
\end{tcolorbox}

The dataset block gives task families and $2{:}2{:}1$ aggregation, but no
held-out instances.  Live state comes from tools rather than self-report.  Near
the context limit, a compaction prompt preserves per-iteration mechanisms,
scores, hypotheses, best snapshot, open questions, and paths; only old messages
are replaced, not disk artifacts.

\subsection{Seed Policy Prompt}
\label{app:seed-policy-prompt}

The minimal seed specifies only CodeAct interaction and workspace boundaries.

\begin{tcolorbox}[
    breakable,
    colback=blue!2,
    colframe=blue!35,
    boxrule=0.5pt,
    arc=1mm,
    left=1.5mm,
    right=1.5mm,
    top=1mm,
    bottom=1mm,
    title={Seed policy system prompt},
    fonttitle=\bfseries\small]
{\footnotesize\ttfamily\raggedright
You are a CodeAct-style task-solving agent.\par
\medskip
You work inside a dedicated task workspace directory. Every
run\_shell\_command runs with that workspace as its working directory, so
relative paths resolve against it. Create, edit, and read your files there.
The concrete absolute workspace path is given in the task message. Use
run\_shell\_command to inspect files, run programs, edit files, and verify your
work. Use finish when the task is done.\par
\medskip
Rules:\par
- Keep all files inside the task workspace.\par
- Do not access paths outside it unless explicitly shown in the task.\par
- Prefer simple shell commands and small file edits.\par
- If a command fails, inspect the output and continue.\par
- Call finish with a concise final answer or artifact path.\par}
\end{tcolorbox}

For task $x$, the seed task analyzer constructs the user message as follows:

\begin{tcolorbox}[
    colback=black!2,
    colframe=black!30,
    boxrule=0.5pt,
    arc=1mm,
    left=1.5mm,
    right=1.5mm,
    top=1mm,
    bottom=1mm]
{\footnotesize\ttfamily\raggedright
Task id: [TASK\_ID]\par
Task workspace (your working directory): [WORKSPACE]\par
\medskip
[PUBLIC TASK PROMPT]\par
\medskip
Public files:\par
- [WORKSPACE-RELATIVE FILE]\par}
\end{tcolorbox}

The file block is optional and the seed omits the domain label.  Scorer fields,
expected outputs, anchors, and rationales are stripped.  Evolved prompt and code
are frozen together.

\subsection{Judge Prompts}
\label{app:judge-prompts}

Judge instructions are fixed and post-rollout.  Deterministic scorers use no
prompt; model judges use Section~\ref{app:model-configurations} and receive
references only inside evaluation (Table~\ref{tab:judge-prompt-design}).

\begin{table}[htbp]
\centering
\small
\caption{Fixed judge-prompt interfaces.  All structured outputs are parsed and
retried on formatting failure; failed grading runs are surfaced as evaluation
errors rather than silently treated as valid model failures.}
\label{tab:judge-prompt-design}
\begin{tabularx}{\textwidth}{p{0.18\textwidth} p{0.34\textwidth} X}
\toprule
\textbf{Scorer} & \textbf{Judge evidence} & \textbf{Required decision} \\
\midrule
BrowseComp / semantic equivalence & Question, protected short reference answer,
and policy final answer & \texttt{CORRECT: YES/NO} plus one-sentence reason;
formatting and equivalent phrasing are ignored. \\

HLE & Question, policy response, and precise protected answer & Extract the
final answer, judge strict equivalence (with a small numerical tolerance), and
return reasoning, a yes/no verdict, and confidence. \\

GDPval absolute & Task, weighted positive/penalty criteria, parsed deliverable
text, formulas, and rendered pages/slides & JSON with points and evidence-backed
reason for every criterion, plus an overall summary. \\

GDPval pairwise & The same rubric plus policy and frozen-seed deliverables,
including text and images & For every criterion choose A, B, or tie.  Both A/B
orders are judged and averaged to reduce position bias. \\

Claw-Eval & Complete executed tool-use trace, task-private grader, rubric, and
reference state & Benchmark-native completion, robustness, communication, and
safety scores, combined by the official grader. \\

APEX-Agents & Policy trajectory, environment-state difference, artifacts, and
task-private verifiers inside the APEX sandbox & Per-verifier rationale and
score; a task passes only when every required criterion passes. \\
\bottomrule
\end{tabularx}
\end{table}

Semantic grading ignores surface form; HLE uses extract-then-compare.  GDPval
uses rendered images for visual criteria and parsed text for facts/formulas;
pairwise grading averages both A/B orders.  APEX-Agents and Claw-Eval retain
native graders.  Validation rationales are evolver-visible but policy-hidden;
held-out feedback is generated only after freezing.

\end{document}